\RequirePackage{fix-cm}
\documentclass[smallextended]{svjour3}
\smartqed
\usepackage{graphicx}
\usepackage{amsmath,amssymb}
\usepackage{color}

\makeatletter
\renewcommand*{\@seccntformat}[1]{\csname the#1\endcsname.\quad}
\makeatother
\newcommand{\argmin}{\mathop{\mathrm{arg\,min}}}
\newcommand{\R}{\mathbb{R}}
\newcommand{\cG}{\mathcal{G}}
\newcommand{\cE}{\mathcal{E}}
\newcommand{\cR}{\mathcal{R}}
\newcommand{\cC}{\mathcal{C}}
\newcommand{\cL}{\mathcal{L}}

\begin{document}

\title{Sequential model aggregation for production forecasting}
\author{Rapha{\"e}l Deswarte \and V{\'e}ronique Gervais \and Gilles Stoltz \and S{\'e}bastien Da Veiga}

\authorrunning{Deswarte, Gervais, Stoltz {\&} Da Veiga}

\institute{Rapha{\"e}l Deswarte \at Ecole Polytechnique
\and V{\'e}ronique Gervais \at IFP Energies nouvelles
\and Gilles Stoltz \at LMO, Universit{\'e} Paris Sud -- CNRS
\and S{\'e}bastien Da Veiga \at IFP Energies nouvelles, now with Safran}

\date{\today}

\maketitle

\thispagestyle{empty}

\renewcommand{\hat}{\widehat}
\renewcommand{\leq}{\leqslant}
\renewcommand{\geq}{\geqslant}
\newcommand{\conv}{\mathop{\mathrm{conv}}}
\renewcommand{\epsilon}{\varepsilon}
\newcommand{\todo}[1]{[{\color{red} #1}]}

\abstract{Production forecasting is a key step to design the future development of a reservoir. A classical way to generate such forecasts consists in simulating future production for numerical models representative of the reservoir. However, identifying such models can be very challenging as they need to be constrained to all available data. In particular, they should reproduce past production data, which requires to solve a complex non-linear inverse problem. In this paper, we thus propose to investigate the potential of machine learning algorithms to predict the future production of a reservoir based on past production data without model calibration. We focus more specifically on robust online aggregation, a deterministic approach that provides a robust framework to make forecasts on a regular basis. This method does not rely on any specific assumption or need for stochastic modeling. Forecasts are first simulated for a set of base reservoir models representing the prior uncertainty, and then combined to predict production at the next time step. The weight associated to each forecast is related to its past performance. Three different algorithms are considered for weight computations: the exponentially weighted average algorithm, ridge regression and the Lasso regression. They are applied on a synthetic reservoir case study, the Brugge case, for sequential predictions. To estimate the potential of development scenarios, production forecasts are needed on long periods of time without intermediary data acquisition. An extension of the deterministic aggregation approach is thus proposed in this paper to provide such multi-step-ahead forecasts.} \enlargethispage{1cm}


\section{Introduction}

Forecasting reservoir production is a key step in the decision-making process for field development. Numerical representations -- or models -- of the reservoirs can be considered to perform such forecasts. They consist of a grid reproducing the structure of the reservoir and populated with facies and petrophysical properties. The evolution of pressure and fluids in the reservoir induced by production is then simulated based on flow equations. The production forecasts should be as reliable as possible to be used for planning the future development of the field. Therefore, they should be constrained to the available observations. The usual way to do this consists in generating such forecasts with reservoir models constrained to all available data. In particular, these models should reproduce the time-dependent measurements acquired during the production period, such as measurements at wells (pressure, oil and water rates, etc.) and 4D-seismic related attributes. As the reservoir model input parameters are not linearly related to these production data, identifying such models requires to solve a non-linear inverse problem, usually referred to as history-matching in reservoir engineering. In practice, solving this problem can be very challenging. Many uncertainties exist, with varying types. They can be related, for instance, to the geological scenario, to the structure of the reservoir, to the spatial distribution of the petrophysical properties, or to the fluid characteristics. In addition, fluid-flow simulations can be very long. Several methods have been investigated to solve the history-matching problem. The variational approach consists in applying minimization algorithms to reduce the error between the production data and the corresponding simulated properties \cite{Tarantola}. Ensemble methods can also be considered, such as the Ensemble Kalman Filter \cite{Aanonsen}. Interested readers can refer to \cite{Oliver}, for instance, for a review of history-matching approaches. \\

Recently, new techniques were proposed in the literature that tackle the problem of reservoir forecasting from a different perspective: forecasts conditioned to past-flow-based measurements are generated without model updating \cite{Satija1,Satija2,Scheidt,Sun2}. These approaches provide uncertainty quantification for the forecasts based on relationships identified between the model-output properties at different times. More precisely, a set of reservoir models is used to represent prior uncertainty. The fluid-flow simulations performed for this ensemble provide a sampling of the data variables and prediction variables, referring to the values simulated for the measured dynamic properties during the history-matching and prediction periods, respectively. The Prediction-Focused Approach (PFA) introduced in \cite{Scheidt} consists in applying a dimensionality reduction technique, namely the non-linear principal component analysis (NLPCA), to the two ensembles of variables (data and prediction). The statistical relationship estimated between the two sets of reduced-order variables is then used to estimate the posterior distribution of the prediction variables constrained to the observations using a Metropolis sampling algorithm. This approach is extended in \cite{Satija1} to Functional Data Analysis. A Canonical Correlation Analysis is also considered to linearize the relationship between the data and prediction variables in the low-dimensional space and then to sample the posterior distribution of prediction variables using simple regression techniques. The resulting approach was demonstrated on a real field case in \cite{Satija2}. In \cite{Sun2}, the data and prediction variables are considered jointly in the Bayesian framework. They are first parameterized using Principal Component Analysis (PCA) combined to some mapping operation that aims at reducing the non-Gaussianity of the reduced-order variables. A randomized maximum likelihood algorithm is then used to sample the distribution of the variables given the observed data.\\

Some machine learning algorithms can also be used to output production predictions over time based on past observations without model updating. As for the techniques described above, these approaches use as building blocks an ensemble of base models, from which forecasts are generated. These base forecasts quantify in some sense some uncertainty (the larger the convex hull of forecasts, the more uncertain). The machine learning algorithms then consist in combining (aggregating) the base forecasts to output predictions. Some of these aggregation techniques deal with stochastic data: the observations to be forecast may be modeled by some stochastic process. On the contrary, other techniques work on deterministic data and come with theoretical guarantees of performance even when the observations cannot be modeled by a stochastic process. Examples of popular aggregation methods include Bayesian model averaging, random forests (stochastic approaches), as well as robust online aggregation (deterministic approach). The latter is also known as prediction of individual sequences or prediction with expert advice \cite{CeLu06}. This sequential aggregation technique, developed in the 1990s, provides a robust framework to make forecasts on a regular (e.g., monthly) basis. It does not rely on any specific assumption or need for stochastic modeling, and was successfully applied for the forecasting of air quality \cite{Oz}, electricity consumption \cite{EDF,EDFBis} and exchange rates \cite{ExchRates}. Here, we propose to assess its performance for reservoir forecasting. The ensemble of base forecasts may be generated from different structural or geological models, as long as they are considered possible representatives of the reservoir. The aggregation approach then provides a production prediction based on the available past production data. This prediction is obtained by combining the base forecasts using convex or linear weights set based on past performance of each base model. This dependency on past performance is where something with a flavor of history-matching is performed. This approach is intrinsically sequential, in the sense that it provides predictions for the next time step only. This will be referred to as one-step-ahead predictions in what follows. In reservoir engineering, production forecasts for a given development scenario are generally required on long periods of time, without intermediary data acquisition. This problem is referred to as the batch case in machine learning. As far as we know, methods dealing with this batch case problem rely on a modeling of the data. In this paper, we thus propose to adapt the robust online aggregation technique to longer-term forecasts, or equivalently multi-step ahead predictions. The basic idea is to apply sequential aggregation to a set of possible sequences of observations during the prediction period in order to obtain some intervals for the production predictions.
\\

The paper outlines as follows. Section~\ref{sec:how} contains a high-level exposition of the machine-learning approach followed (with some additional practical details on the implementation of the algorithms being provided later in Section~\ref{sec:faisceaux}). Section~\ref{sec:1} describes the synthetic data set used -- the Brugge case. Section~\ref{sec:ponctuel} discusses our one-step-ahead predictions for this test case while Section~\ref{sec:faisceaux} shows our longer-term predictions.

\section{How to combine production forecasts}
\label{sec:how}

Our approach aims to predict the value of a given property $y$ forward in time based on past observations of this property. In the case of reservoir engineering, property $y$ stands for production data such as pressure and oil rate at a given well, or cumulative oil produced in the reservoir.

In what follows, we assume that a sequence of observed values for $y$ at times $1,\ldots,T-1$ is available, and we denote it by $y_1,y_2,\ldots,y_{T-1}$. The time interval between two consecutive observations is assumed to be regular. Then the proposed approach provides prediction for property $y$ at subsequent times $T, \ldots, T+K$ following the same time frequency as the observations. To do so, a set of $N$ reservoir models representative of the prior uncertainty is generated, and fluid-flow simulations are performed to obtain the values of $y$ at times $t=1  \ldots T+K$. These simulated values will be denoted by $m_{j,t}$, where $j = 1 \ldots N$ indexes the models. To estimate property $y$ at time $T$, we propose to apply existing aggregation algorithms ``from the book'', with no tweak or adjustment that would be specific to the case of reservoir production (section \ref{sec:highlevel1}). An extension of these techniques to predict longer-term forecasts, i.e., at times $T+1$,...,$T+K$, is then proposed in section \ref{sec:highlevel2}.

\subsection{High-level methodology: point aggregation for one-step-ahead forecasts}
\label{sec:highlevel1}

The aggregated forecast $\hat{y}_T$ at time $T$ is obtained by linearly combining the simulated forecasts $m_{j,T}$:
\begin{equation}
\label{eq:aggreg}
\hat{y}_T = \sum_{j=1}^{N} w_{j,T} \, m_{j,T}\,,
\end{equation}
where the weights $w_{j,T}$ are determined based on the past
observations $y_t$ and past forecasts $m_{j,t}$, where $t \leq T-1$.
The precise formulae to set these weights (some specific algorithms
designed by the literature) are detailed in Section~\ref{sec:appalgo} below.
The basic idea is to put higher weights on models that performed
better in the past.

The main interest of this methodology is given by its performance guarantee:
the weights can be set to mimic the performance of some good constant
combination of the forecasts. More precisely, given a set $\mathcal{W}^\star$
of reference weights (e.g., the set of all convex weights, or of all linear weights in
some compact ball), good algorithms ensure that,
no matter what the observations $y_t$ and the forecasts $m_{j,t}$ of the models were,
\begin{equation}
\label{eq:theoguar}
\frac{1}{T} \sum_{t=1}^T \bigl( \hat{y}_t - y_t \bigr)^2
\leq \epsilon_T + \inf_{(v_1,\ldots,v_{N}) \in \mathcal{W}^\star}
\frac{1}{T} \sum_{t=1}^T \left( \hat{y}_t - \sum_{j=1}^{N} v_j \, m_{j,t} \right)^{\!\! 2}\,,
\end{equation}
where $\epsilon_T$ is a small term, typically of order $1/\sqrt{T}$.
More details are given in Section~\ref{sec:appalgo}, for each specific algorithm.

The reference set $\mathcal{W}^\star$ will always include
the weights $(v_1,\ldots,v_{N})$ of the form $(0,\ldots,0,1,0,\ldots,0)$
that only put non-zero mass equal to $1$ on one model.
Thus, the infimum over elements in $\mathcal{W}^\star$ will always be smaller
than the cumulative square loss of the best of the $N$ models.
For some algorithms, this reference set $\mathcal{W}^\star$ will be much larger
and will contain all weights of some Euclidean ball of $\R^{N}$ with radius larger
than~$1$, thus in particular, all convex weights.

The algorithms we will consider (and the way we will refer to them in the sequel) are:
the exponentially weighted average (EWA) forecaster;
the ridge regression (Ridge);
the Lasso regression (Lasso). Their statement and theoretical guarantees -- in
terms of the quantities $\mathcal{W}^\star$ and $\varepsilon_T$ in~\eqref{eq:theoguar} -- are
detailed in Section~\ref{sec:appalgo}.

Before we describe them in details, we provide a high-level view
on the second aspect of our methodology, pertaining to
longer-term predictions.

\subsection{High-level methodology: multi-step-ahead predictions}
\label{sec:highlevel2}

As previously, we assume here that only observations until time $T-1$ are available. The aggregated forecast at time $T$ can be obtained using the one-step-ahead techniques described above. In this section, we focus on the prediction of longer-term forecasts, i.e., for rounds $t = T+k$, where $k \geq 1$
and $k$ can be possibly large.

To predict such multi-step-ahead forecasts, we propose to apply point aggregation methods on a set of plausible values of the measurements at times $T,\ldots,T+K$.
For each $k = 1,\ldots,K$, this approach thus provides at each round $T+k$ a set $\hat{\mathcal{W}}_{T+k}$ of possible weights $(w_{1,T+k},\ldots,w_{N,T+k})$. The interval forecast for round $t = T+k$ is then
\begin{equation}
\label{eq:computational}
\hat{S}_{T+k}
= \conv \! \left\{ \sum_{j=1}^{N} w_{j,T+k} \, m_{j,T+k} : \ \
(w_{1,T+k},\ldots,w_{N,T+k}) \in \hat{\mathcal{W}}_{T+k}
\right\},
\end{equation}
where $\conv$ denotes a convex hull, possibly with some enlargement to take into account
the noise level.

\begin{figure}
\includegraphics[width=0.43\textwidth]{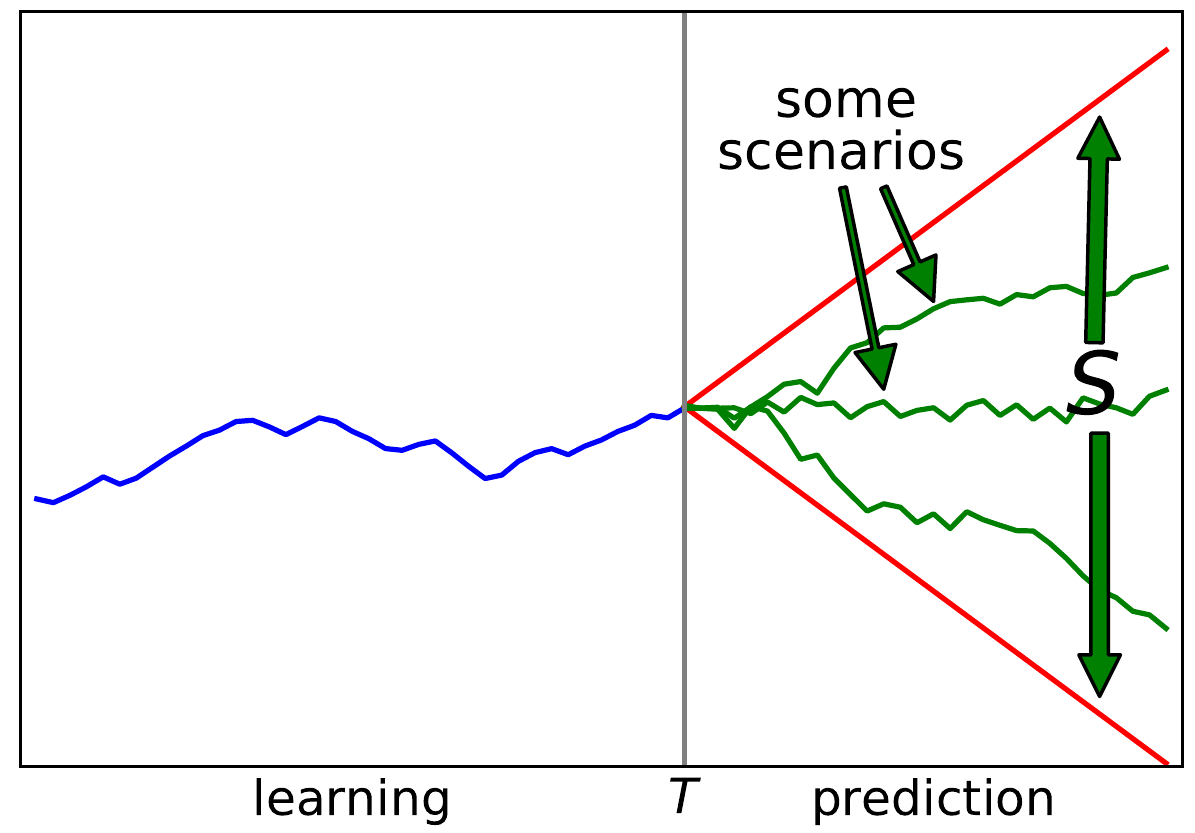}
\hfill
\includegraphics[width=0.43\textwidth]{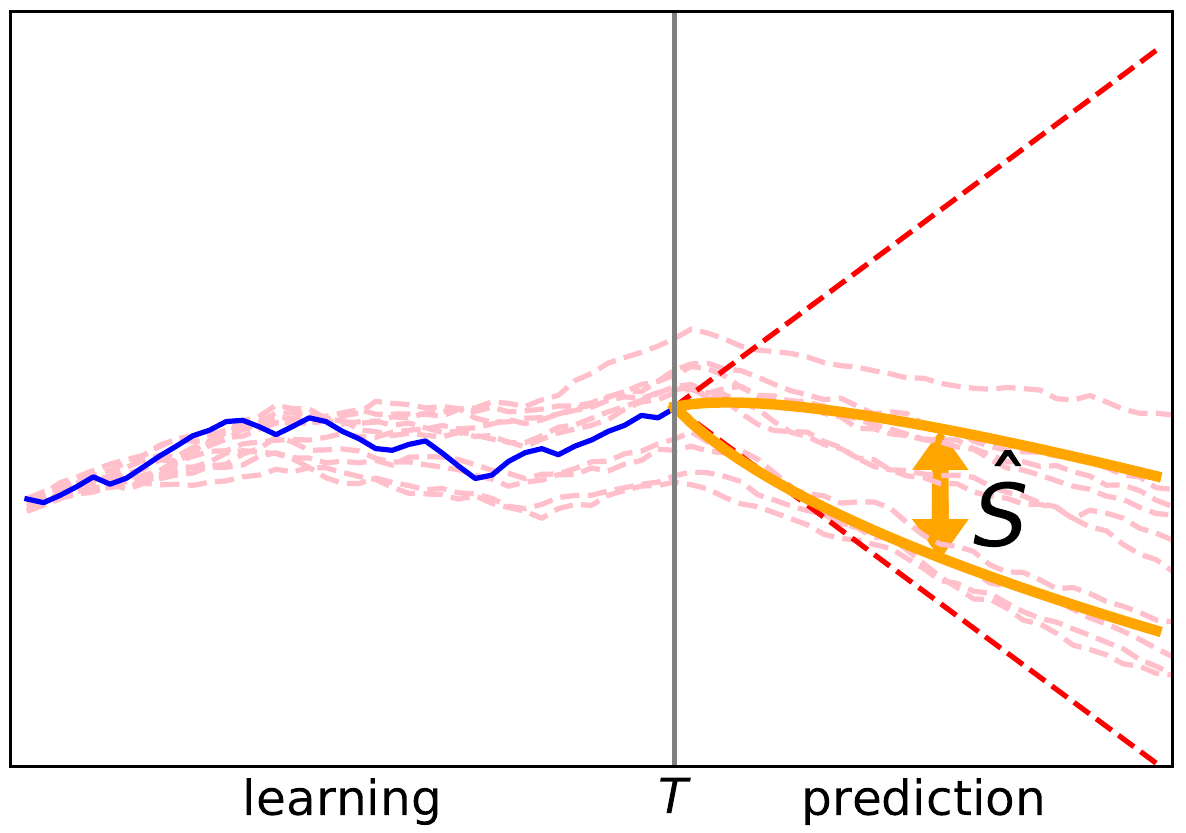}
\caption{\label{fig:interval-principle} Schematic diagram for multi-step-ahead predictions.}
\end{figure}

More precisely, the approach encompasses the following steps, also illustrated in Figure~\ref{fig:interval-principle}:

\begin{enumerate}
\item we consider the set $S = S_{ T} \times \ldots \times S_{T+K}$ of
all plausible continuations  $z_{T},\ldots,z_{T+K}$ of the observations $y_1,\ldots,y_{T-1}$; this
set $S$ will be referred to as the set of scenarios;
\item for each given scenario $y_1,\ldots,y_{T-1},z_{T},\ldots,z_{T+K} \in S$,
\begin{itemize}
\item for each round $T+k$, where $k = 1,\ldots,K$,
we compute the weights $(w_{1,T+k},\ldots,w_{N,T+k})$ corresponding to the putative
past observations $y_1,\ldots,y_{T-1},z_{T},\ldots,z_{T+k-1}$
and corresponding model forecasts;
\item we form the aggregated forecast $\hat{z}_{T+k} = \sum_{j} w_{j,T+k} \, m_{j,T+k}$.
\end{itemize}
\item
The interval forecasts $\hat{S}_{T+k}$ are the convex hulls of all possible aggregated
forecasts $\hat{z}_{T+k}$ obtained by running all scenarios in $S$
(possibly with some enlargement to take into account the noise level
and with some initial matching at time $T$).
\end{enumerate}

The main constructions remaining to be explained is
(i) how the set $S$ of plausible continuations
is determined; (ii) how we may efficiently compute the interval forecasts $\hat{S}_{T+k}$,
as there are infinitely many scenarios;
what we mean by (iii) an enlargement to account for the noise level
and (iv) an initial matching at time $T$.

\begin{figure}
\includegraphics[width=0.48\textwidth]{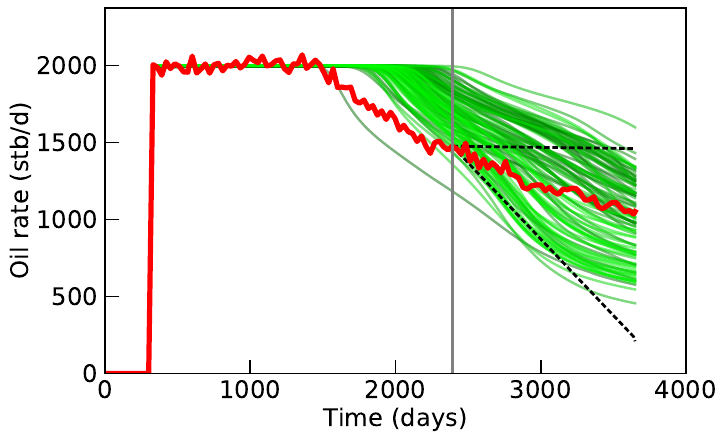}
\hfill
\includegraphics[width=0.4\textwidth]{Fig/Methodo/Faisc-L.pdf}
\caption{\label{fig:cone-principle} An example of the set of scenarios $S$
calculated for oil rate at a producer (QO\_P19, see Section~\ref{sec:1}) on the last third of the simulation period (left) and a repetition of the
corresponding schematic diagram (right). On the left figure, production data are plotted in red and simulated forecasts in green. }
\end{figure}

\paragraph{Set of plausible continuations.}
We propose here an assisted way to build the set $S$ of possible scenarios~(i). The high-level idea is that we look on available data how large the typical
variations were. More precisely, we compute the maximal variations upwards $M$ or downwards $m$ of the observations
on the learning part of the data set (times $1$ to $T-1$), and of any single trajectory of model forecasts on the
prediction part of the data set (times $T$ to $T+K$). We do so by considering variations averaged out over 10 consecutive steps.
The maximal variation downwards $m$ can be negative or positive, depending on the
property considered; the same remark holds for the maximal variation upwards.
This yields an interval $[m,M]$ of typical 1-step average
variations. The set of scenarios is then the cone
formed by the product of the intervals $[y_{T-1} + km, \, y_{T-1} + kM]$, where
$k = 0, 1, 2, \ldots$. See Figure~\ref{fig:cone-principle} for an illustration. This approach provides a first guess for $S$, that can be adjusted afterwards depending on the specificity of the problem. For instance, physical bounds can be introduced to constrain $S$, such as  maximum values for the oil rates.

\paragraph{Efficient computation of the interval forecasts.}
As for (ii), a computational issue is to be discussed: getting a numerical value of the
convex hull~\eqref{eq:computational} is computationally challenging
as $S$ typically contains infinitely many scenarios. However, we should note that only
the upper and lower bounds of $\hat{S}_{T+k}$ need to be computed (or bounded). We could provide
a solution to this problem for two of the considered algorithms, namely Ridge and EWA. As the corresponding formulae are however not easy to describe, we prefer
to omit them and refer the interested reader to \cite[Chapter~5]{Desw17}.
For Ridge, we were able therein to determine a closed-form expression of the upper and lower
bounds of the sets $\hat{S}_{T+k}$ in~\eqref{eq:computational}. As for EWA, we offer therein an efficient and recursive computation of a series of sets
$\smash{\hat{\mathcal{W}}'_{T+k}}$ containing the target sets $\smash{\hat{\mathcal{W}}_{T+k}}$,
from which it is then easy to compute intervals containing the target
prediction intervals $\hat{S}_{T+k}$. Indeed, it suffices to compute
the maximum and the minimum of each~$\hat{S}_{T+k}$.

\paragraph{Enlargement of the interval forecasts to take noise into account.}
We first study the learning part of the data set and
estimate some upper bound $\sigma_{\max}$ on the noise level of the observations,
as detailed below.
Then, denoting by $c_{T+k}$ the center of each interval forecast $\hat{S}_{T+k}$,
we replace
\[
\hat{S}_{T+k}
\qquad \mbox{by} \qquad
\max \Bigl\{ \hat{S}_{T+k}, \,\, [c_{T+k}-\sigma_{\max},\,c_{T+k}+\sigma_{\max}]
\Bigr\}\,,
\]
where the maximum in the right-hand side has to be understood in terms
of the inclusion $\subseteq$ operator.

Our estimate $\sigma_{\max}$ is formed as follows, based on an interpretation
of noise as some intrinsic variability.
We first determine, among the observations available,
time steps corresponding to some local stability of the property studied;
those steps $t$ are the ones when the observation $y_t$ is within $150$ psi or
$150$ bbl/day (depending of the property) of all $y_{t-r}$,
where $r$ varies between $-15$ and $+15$.
We denote by $\mathcal{S}$ the set of those time steps with local stability.
Then, our estimate is
\[
\sigma_{\max} = \max_{t \in \mathcal{S}}
\left| y_t - \frac{1}{5}\sum_{r=t-2}^{t+2} y_r \right|.
\]

\paragraph{Initial matching at $T$.}
All algorithms make some prediction error when
forecasting $y_T$ by $\hat{y}_T$ at the beginning of the prediction period. To avoid that this error of $\Delta_T = \hat{y}_T - y_T$
be carried over the whole prediction part of the data set (over
all time steps $T+k$, where $k \geq 1$), we shift all interval forecasts
$\hat{S}_{T+k}$ by this initial error $\Delta_T $ (in the case of EWA)
or by the average of $\Delta_{T-4},\ldots,\Delta_T$ (in the case of Ridge).

\subsection{Statement of the point aggregation algorithms considered}
\label{sec:appalgo}

We provided in Section~\ref{sec:highlevel1} the general methodological framework
for sequentially aggregating forecasts in a robust manner, not relying on any
stochastic modeling of the observations or of the forecasts of the models.
We now provide the statements and the theoretical
guarantees of the considered algorithms; the theoretical guarantees
refer to~\eqref{eq:theoguar} and consist in providing the values of $\mathcal{W}^\star$
and $\varepsilon_T$ for the considered algorithm.

\subsubsection{The ridge regression (Ridge)}

The ridge regression (which we will refer to as Ridge when reporting the experimental results)
was introduced by~\cite{HoKe70} in a stochastic and non-sequential setting.
What follows relies on recent new analyses of the ridge regression
in the machine learning community; see the original papers by~\cite{Vov01,AzWa01}
and the survey in the monograph by~\cite{CeLu06}, as well as the discussion and the optimization
of the bounds found in these references proposed by~\cite{Gerchino}.

Ridge relies on a parameter $\lambda > 0$, called a regularization factor.
At round $T = 1$, it picks arbitrary weights, e.g., uniform $(1/N,\ldots,1/N)$ weights.
At rounds $T \geq 2$, it picks
\begin{equation}
\label{eq:defridge}
(w_{1,T},\ldots,w_{N,T}) \in \argmin_{(v_1,\ldots,v_{N}) \in \R^{N}}
\left\{ \lambda \sum_{j=1}^{N} v_j^2 +
\sum_{t=1}^{T-1} \Biggl( y_t - \sum_{j=1}^{N} v_j \, m_{j,t} \Biggr)^{\!\! 2}
\right\};
\end{equation}
i.e., it picks the best constant weights to reconstruct past observations based on the
model forecasts subject to an $\ell^2$--regularization constraint $\sum v_j^2$, which
is useful to avoid overfitting to the past.

The performance bound relies on two bounds $V$ and $B$ and is over
\[
\mathcal{W}^\star = \left\{ (v_1,\ldots,v_N) : \ \ \sum_{j=1}^N v_j^2 \leq V^2 \right\},
\]
the Euclidean ball of $\R^N$ with center $(0,\ldots,0)$ and radius $V \geq 1$.
This ball contains in particular all convex combinations in $\R^N$.
The bound~\eqref{eq:theoguar} with the above $\mathcal{W}^\star$ reads:
for all bounded sequences of observations $y_t \in [-B,B]$ and model forecasts $m_{j,t} \in [-B,B]$, where $t=1 \ldots T$,
\[
\varepsilon_T \leq \frac{1}{T} \left(
\lambda V^2 + 4 N B^2 \biggl( 1+ \frac{N B^2 T}{\lambda} \biggr)
\ln \biggl( 1 + \frac{B^2 T}{\lambda} \biggr) + 5 B^2
\right).
\]
In particular, for a well-chosen $\lambda$ of order $\sqrt{T}$, we have
$\varepsilon_T = O\bigl((\ln T) / \sqrt{T} \bigr)$.

The latter choice on $\lambda$ depends however on the quantities $T$ and $B$,
which are not always known in advance. This is why in practice we set the $\lambda$
to be used at round $t$ based on past data.
More explanations and details are provided in Section~\ref{sec:implement} below.

\subsubsection{The Lasso regression (Lasso)}

The Lasso regression was introduced by~\cite{Tib96}, see also the
efficient implementation proposed in~\cite{LARS}. Its definition is similar to the
definition~\eqref{eq:defridge} of Ridge, except that the $\ell^2$--regularization is replaced
by an $\ell^1$--regularization: at rounds $T \geq 2$,
\[
(w_{1,T},\ldots,w_{N,T}) \in \argmin_{(v_1,\ldots,v_{N}) \in \R^{N}}
\left\{ \lambda \sum_{j=1}^{N} |v_j| +
\sum_{t=1}^{T-1} \Biggl( y_t - \sum_{j=1}^{N} v_j \, m_{j,t} \Biggr)^{\!\! 2}
\right\}.
\]
As can be seen from this definition,
Lasso also relies on a regularization parameter $\lambda > 0$.

One of the key features of Lasso is that the weights $(w_{1,T},\ldots,w_{N,T})$
it picks are often sparse: many of its components are null.
Unfortunately, we are not aware of any performance guarantee of the form~\eqref{eq:theoguar}:
all analyses of Lasso we know of rely on (heavy) stochastic assumptions and
are tailored to non-sequential data. We nonetheless implemented it and tabulated its
performance. \\

Another appealing regularized regression is the elastic net introduced by~\cite{ZouHastie05}:
it seeks the best compromise between Lasso and Ridge by considering both regularizations,
i.e., by adding a
\[
\lambda_1 \sum_{j=1}^{N} |v_j| + \lambda_2 \sum_{j=1}^{N} v_j^2
\]
regularization to the least-square criterion, where $\lambda_1,\,\lambda_2$
are two parameters to be tuned. Its known analyses are also only tailored to
non-sequential data. The method would be subject of study in future
developments of the present approach. 

\subsubsection{The exponentially weighted average (EWA) forecaster}

The previous two forecasters were using linear weights: weights that lie in $\R^N$
but are not constrained to be nonnegative or to sum up to~$1$. In contrast, the
exponentially weighted average (EWA) forecaster picks convex weights: weights that
are nonnegative and sum up to~$1$. The aggregated forecast $\hat{y}_T$
lies therefore in the convex hull of the forecasts $m_{j,T}$ of the models,
which may be considered a safer way to predict.

EWA (sometimes called Hedge)
was introduced by \cite{Vov90,LiWa94} and further understood and studied by, among others,
\cite{CeFrHaHeScWa97,Ces99,AuCeGe02}; see also the monograph by~\cite{CeLu06}.
This algorithm picks uniform $(1/N,\ldots,1/N)$ weights at round $T = 1$, while at subsequent
rounds $T \geq 2$, it picks weights $(w_{1,T},\ldots,w_{N,T})$ such that
\[
w_{j,T} = \frac{\displaystyle{\exp \! \left( - \eta \sum_{t=1}^{T-1} (y_t - m_{j,t})^2 \right)}}{
\displaystyle{\sum_{k=1}^N \exp \! \left( - \eta \sum_{t=1}^{T-1} (y_t - m_{k,t})^2 \right)}
}\,.
\]
The weight put on model $j$ at round $T$ depends on the cumulative accuracy error
suffered by $j$ on rounds $1$ to $T-1$; however, the weight is not directly proportional
to this cumulative error: a rescaling via the exponential function is operated, with a
parameter $\eta > 0$. We will call this parameter the learning rate of EWA: when $\eta$
is smaller, the weights get closer to the uniform weights; when $\eta$ is larger,
the weights of the suboptimal models get closer to~$0$ while the (sum of the) weight(s) of the
best-performing model(s) on the past get closer to~$1$.

To provide the performance bound we first denote by $\delta_j$
the convex weight vector $(0,\ldots,0,1,0,\ldots,0)$, where the unique
non-zero coordinate is the $j$--th one. The set $\mathcal{W}^\star$
of reference weights is given by
\[
\mathcal{W}^\star = \bigl\{ \delta_j : \ \ j \in \{1,\ldots,N\} \bigr\}\,.
\]
The performance bound~\eqref{eq:theoguar} with the above $\mathcal{W}^\star$ relies
on a boundedness parameter $B$ and reads:
for all bounded sequences of observations $y_t \in [0,B]$ and model forecasts $m_{j,t} \in [0,B]$,
\[
\varepsilon_T \leq
\left\{
\begin{array}{ccc}
\displaystyle{\frac{\ln N}{\eta T}} \phantom{+ \frac{\eta B^2}{8}} &
\phantom{espa} & \mbox{if} \ \eta \leq 1/(2B^2), \vspace{.2cm}\\
\displaystyle{\frac{\ln N}{\eta T}} + \frac{\eta B^2}{8} & & \mbox{if} \ \eta > 1/(2B^2). \\
\end{array}
\right.
\]
In particular, $\epsilon_T = O(1/T)$ if $\eta$ is well-calibrated,
which requires the knowledge of a plausible bound $B$. Here again, we may prefer
to set the $\eta$ to be used at round $t$ based on past data;
see Section~\ref{sec:implement} below.

\subsubsection{How to implement these algorithms (i.e.,
pick their parameter $\lambda$ or $\eta$)}
\label{sec:implement}

First, note that the algorithms described above rely each on a single
parameter $\lambda > 0$ or $\eta > 0$, which is much less than the number of parameters to be tuned in reservoir models during history-matching. (These parameters
$\lambda$ and $\eta$ are actually rather called hyperparameters to
distinguish them from the model parameters.)

In addition, the literature provides
theoretical or practical guidelines on how to choose these parameters.
The key idea was introduced by~\cite{AuCeGe02}. It consists in letting the parameters $\eta$ or
$\lambda$ vary over time: we denote by $\eta_t$ and $\lambda_t$ the parameters
used to pick the weights $(w_{1,t},\ldots,w_{N,t})$ at round~$t$. Theoretical
studies offer some formulae for $\eta_t$ and $\lambda_t$ (see, e.g.,
\cite{AuCeGe02,SOE}) but the associated practical performance are usually poor,
or at least, improvable, as noted first by \cite{EDF} and later by \cite{ExchRates}.
This is why \cite{EDF} suggested and implemented the following tuning of $\eta_t$ and $\lambda_t$ on past data,
which somehow adapts to the data without overfitting; it corresponds to a grid search of the
best parameters on available past data.

More precisely, we respectively denote by $\cR_\lambda$, $\cL_\lambda$ and $\cE_\eta$ the algorithms Ridge and Lasso
run with constant regularization factor $\lambda > 0$ and EWA run with constant learning
rate $\eta > 0$. We further denote by $L_{T-1}$ the cumulative loss they suffered
on prediction steps $1,2,\ldots,T-1$:
\[
L_{T-1}(\,\cdot\,) = \sum_{t=1}^{T-1} \bigl( \hat{y}_t - y_t)^2\,,
\]
where the $\hat{y}_t$ denote the predictions output by the algorithm
considered, $\cR_\lambda$, $\cL_\lambda$ or $\cE_\eta$. Now, given a finite grid $\cG \subset (0,+\infty)$
of possible values for the parameters $\lambda$ or $\eta$, we pick, at round $T \geq 2$,
\begin{align*}
\lambda_T \in \argmin_{\lambda \in \cG} \bigl\{ L_{T-1}(\cR_\lambda) \bigr\}\,,
\qquad &
\lambda_T \in \argmin_{\lambda \in \cG} \bigl\{ L_{T-1}(\cL_\lambda) \bigr\} \\
\qquad \mbox{and} \qquad &
\eta_T \in \argmin_{\eta \in \cG} \bigl\{ L_{T-1}(\cE_\eta) \bigr\}\,,
\end{align*}
and then form our aggregated prediction $\hat{y}_T$ for step~$T$ by using either the aggregated forecast output by $\cR_{ \lambda_T}$,
$\cL_{ \lambda_T}$, or $\cE_{ \lambda_T}$.

We resorted to wide grids in our implementations, as the various properties to
be forecast have extremely different orders of magnitude:
\begin{itemize}
\item for EWA, 300 equally spaced in logarithmic scale between $10^{-20}$ and $10^{10}$;
\item for Lasso, 100 such points between $10^{-20}$ and $10^{10}$;
\item for Ridge, 100 such between $10^{-30}$ and $10^{30}$.
\end{itemize}
However, how fine the grids are has not a significant impact on the performance;
what matters most is that the correct orders of magnitude for the
hyperparameters be covered by the considered grids.

\section{Reservoir case study}
\label{sec:1}

\begin{figure}
\includegraphics{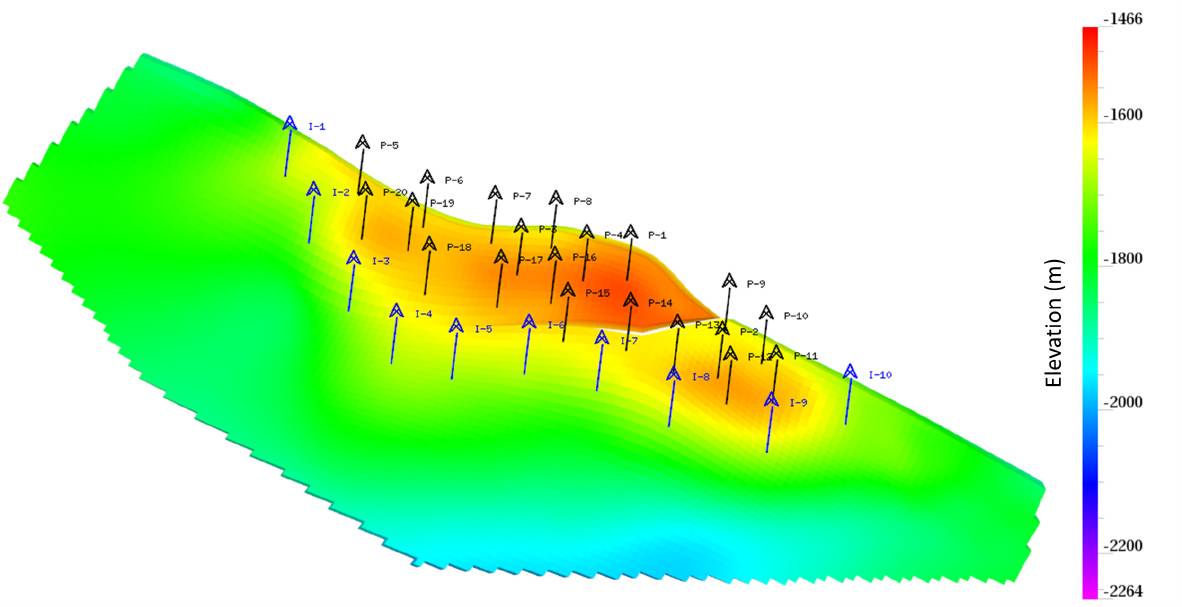}
\caption{\label{fig:1}
Structure of the Brugge field and well location.
Producers are indicated in black and injectors in blue.}
\end{figure}

We consider here the Brugge case, defined by TNO for benchmark purposes \cite{brugge}, to assess the potential of the proposed approach for reservoir engineering. This field, inspired by North Sea Brent-type reservoirs, has an elongated half-dome structure with a modest throw internal fault as shown in Figure \ref{fig:1}. Its dimensions are about 10km $\times$ 3km. It consists of four main reservoir zones, namely Schelde, Waal, Maas and Schie. The formations with good reservoir properties, Schelde and Maas, alternate with less permeable regions. The average values of the petrophysical properties in each formation are given in Table \ref{tab:1}. The reservoir is produced by 20 wells located in the top part of the structure. They are indicated in black in Figure \ref{fig:1} and denoted by $Pj$, with $j = 1, \ldots, 20$. Ten water injectors are also considered. They are distributed around the producers, near the water-oil contact (blue wells in Figure \ref{fig:1}), and are denoted by $Ii$, where $i = 1, \ldots, 10$. \\

\begin{table}
\caption{\label{tab:1}
Petrophysical properties of the Brugge field}
\begin{tabular}{llllll}
\hline\noalign{\smallskip}
Formation & Depositional  & Average  & Average & Average  & Average \\
 & environment & thickness (m) & Porosity & Permeability (mD) & Net to Gross\\
\noalign{\smallskip}\hline\noalign{\smallskip}
Schelde & Fluvial & 10 & 0.207 & 1105 & 60  \\
Waal & Lower shoreface & 26 & 0.19 & 90 & 88\\
Maas & Upper shoreface & 20 & 0.241 & 814 & 97 \\
Schie & Sandy shelf & 5 & 0.194 & 36 & 77\\
\noalign{\smallskip}\hline
\end{tabular}
\end{table}

A fine-scale reference geological model of 20 million grid blocks was initially generated and populated with properties by TNO. It was then upscaled to a 450\,000 grid block model used to perform the fluid-flow simulation considered in the following as the reference one. The reservoir is initially produced under primary depletion. The producing wells are opened successively during the first 20 months of production. They are imposed a target production rate of 2\,000 bbl/day, with a minimum bottomhole pressure of 725 psi. Injection starts in a second step, once all producers are opened. A target water injection rate of 4\,000 bbl/day is imposed to the injectors, with a maximal bottomhole pressure of 2\,611 psi. A water-cut constraint of 90\% is also considered at the producers.\\

The aim of our study is to apply the one-step-ahead and multi-step-ahead approaches described in Section~\ref{sec:how} to predict the production data provided by TNO, namely the oil and water rates at the producers and bottomhole pressure at all wells (see Table~\ref{tab:summary}). To do so, we consider the set of 104 geological models of 139 $\times$ 48 $\times$ 9 grid blocks ($\sim$ 60\,000) provided to the project participants to represent the prior geological uncertainty. These models were built considering various approaches for the simulation of facies, fluvial reservoir zones, porosity and permeability, and are conditioned to static data extracted from the reference case. They differ in the distributions of facies, porosity, net-to-gross, water saturation and permeability. More details can be found in \cite{brugge}, together with examples of permeability realizations.

\begin{figure}
\begin{center}
\includegraphics[width=\textwidth]{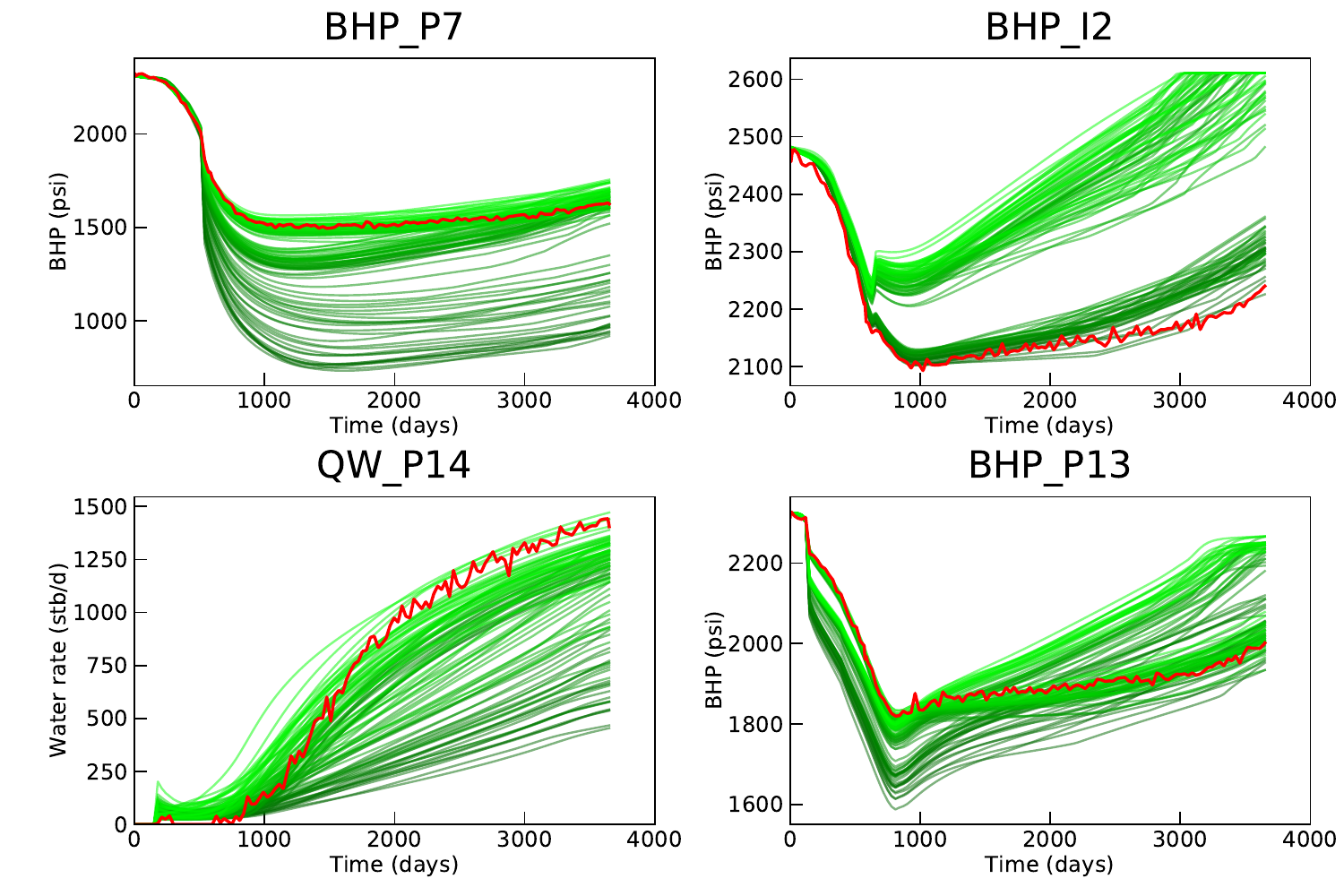} \\
\includegraphics[width=\textwidth]{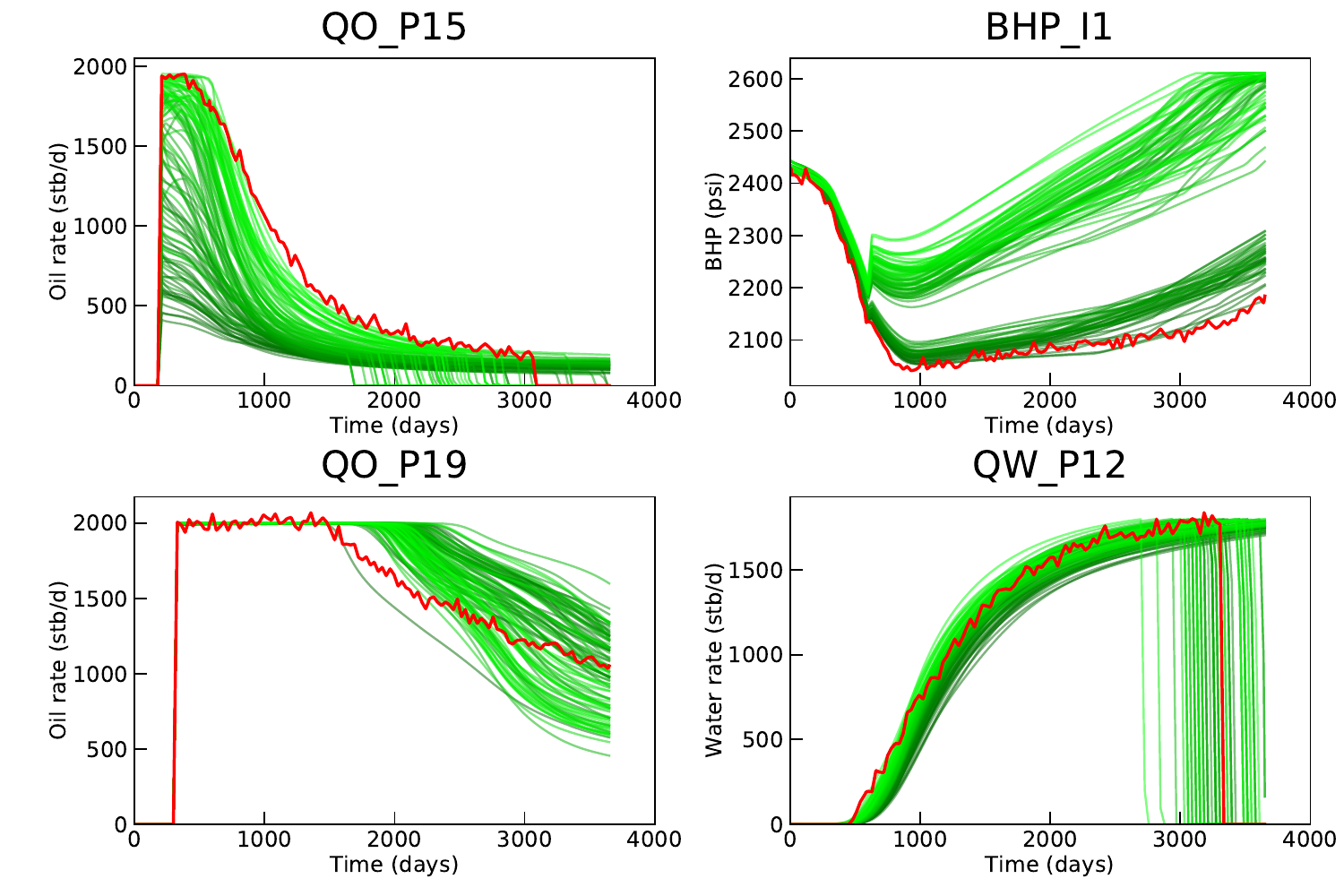}
\end{center}
\caption{\label{fig:uncertainty} Reference production data [red] versus
values simulated with the 104 models [green], over time (in days).}
\end{figure}

Two-phase flow simulations were performed for each of the $104$ reservoir models, considering the same fluid and well properties. Production constraints are the ones used for the reference case. This provides a set of $104$ base forecasts for the $70$ production time-series to be predicted. Some of them are shown in Figure~\ref{fig:uncertainty}.\\

The available data cover 10 years of production. This time interval was split here into 127 evenly spaced prediction steps, which are thus roughly separated by a month. The aggregation algorithms were first applied to sequentially predict the production at wells on this (roughly) monthly basis (Section~\ref{sec:ponctuel}). Then, the approach developed for longer-term forecasts was considered to estimate an interval forecast for the last 3.5 years of production (Section~\ref{sec:faisceaux}). In both cases, the approaches were applied to each time-series independently, i.e., property by property and well by well.\\

Table~\ref{tab:summary2} provides some descriptive statistics pertaining to
the orders of magnitude of the time-series. They should be put in perspective
with the root mean-square errors calculated later in Section~\ref{sec:RMSE}.
In this table, we report both descriptive statistics for the
original (nominal) time-series, as well as for the time-series of
unit changes\footnote{We suppressed the extreme changes
caused by an opening or a closing of the well when
computing the mean, the median, and the standard deviation
of the absolute values of these changes.}
(variations between two prediction steps).
The latter are the most interesting ones in our view,
as far as one-step-ahead forecasting is concerned.

\begin{table}
\caption{\label{tab:summary}
Summary of the 70 times-series to be predicted.}
\begin{tabular}{llllr}
\hline\noalign{\smallskip}
Notation & Well type & Property measured & Units & Number \\
\noalign{\smallskip}\hline\noalign{\smallskip}
BHP\_I$i$ & injector & bottomhole pressure & psi & 10 \\
BHP\_P$j$ & producer & bottomhole pressure & psi & 20 \\
QO\_P$j$ & producer & flow rate of oil & bbl/day & 20 \\
QW\_P$j$ & producer & flow rate of water & bbl/day & 20 \\
\noalign{\smallskip}\hline
\end{tabular}
\end{table}

\begin{table}
\caption{\label{tab:summary2}
Some descriptive statistics on the properties to be predicted,
by type of property. The upper part of the table discusses
the nominal time-series, while its lower part studies
the time-series of the unit changes.}
\begin{tabular}{lrrrr}
\hline\noalign{\smallskip}
Property & BHP\_I & BHP\_P & QO & QW \\
Units & psi & psi & bbl/day & bbl/day \smallskip \\
\hline\noalign{\smallskip}
\multicolumn{5}{l}{Observations} \\
-- Minimum & 2\,007 & 708 & 0 & 0 \\
-- Maximum & 2\,488 & 2\,380 & 2\,147 & 1\,870 \\
\multicolumn{5}{l}{Forecasts output by the models} \\
-- Minimum & 1\,973  & 723 & 0 & 0 \\
-- Maximum & 2\,610  & 2\,443 & 2\,002 & 1\,800 \\
\noalign{\smallskip}
\hline\noalign{\smallskip}
\multicolumn{5}{l}{Unit changes} \\
-- Minimum & $-52$ & $-1\,308$ & $-306$ & $1\,824$ \\
-- Maximum & 39  & 487 & 2\,147 & 156 \\
\multicolumn{5}{l}{Absolute values of the unit changes} \\
-- Mean & 10 & 12 & 32 & 18\\
-- Median & 8 & 9 & 25 & 0 \\
-- Standard deviation & 8 & 13 & 28 & 28\\
\noalign{\smallskip}\hline
\end{tabular}
\end{table}

\definecolor{purp}{rgb}{0.8, 0, 0.8}
\definecolor{grey}{rgb}{0.5, 0.5, 0.4}
\definecolor{pink}{rgb}{0.96, 0.73, 1.0}
\definecolor{strongyellow}{rgb}{0.8, 0.8, 0}

\section{Results of point aggregation for one-step-ahead forecasts}
\label{sec:ponctuel}

We discuss here the application of the Ridge, Lasso and EWA algorithms
on the Brugge data set for one-step-ahead predictions, which (roughly) correspond to one-month-ahead predictions.

\subsection{Qualitative study}

\begin{figure}
\begin{center}
\includegraphics[width=\textwidth]{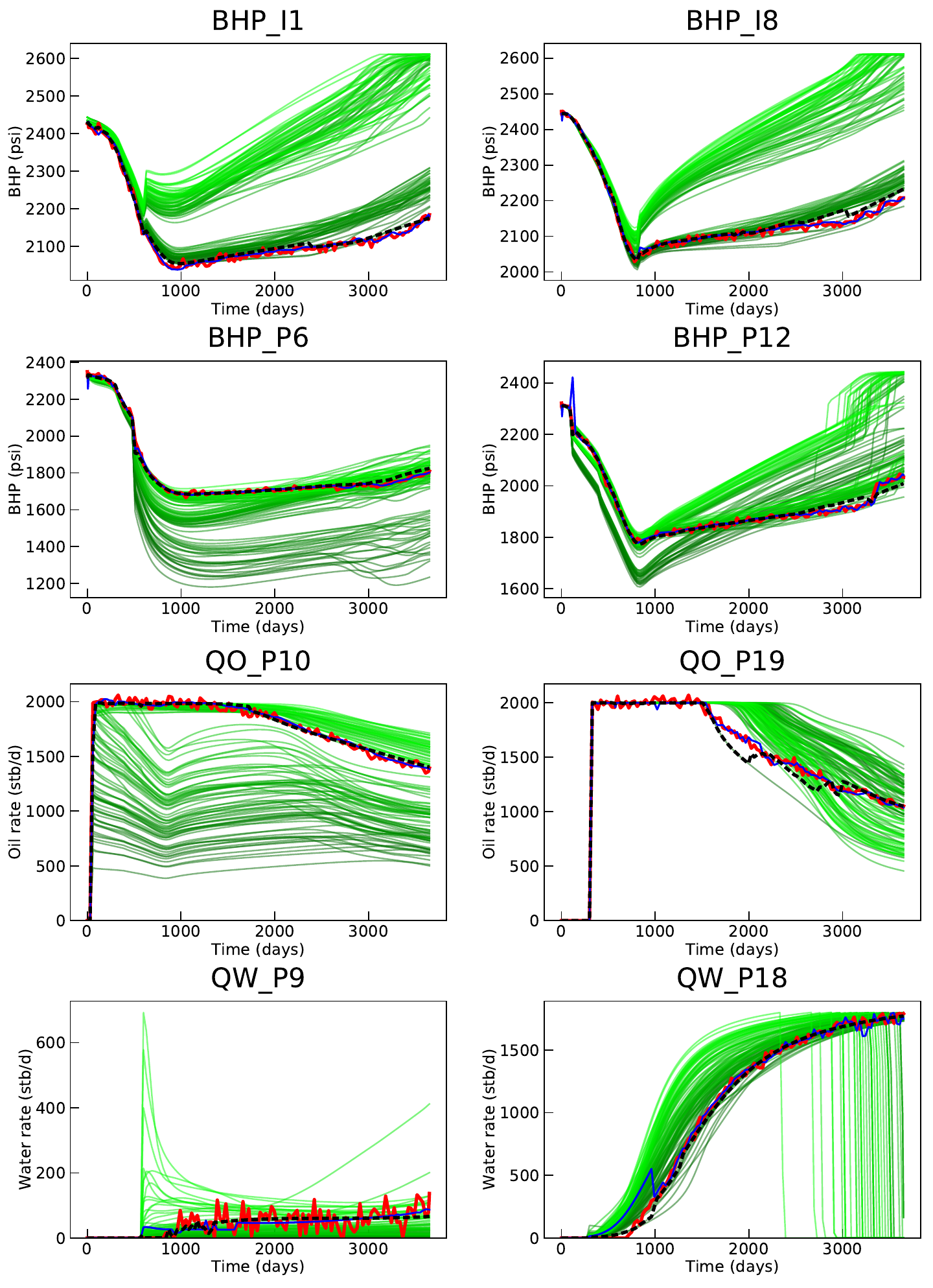}
\end{center}
\caption{\label{fig:RMSE-quali} Values simulated for the 104 models ({\color{green} green lines ---}),
observations ({\color{red} red solid line~---}),
one-step-ahead
forecasts by Ridge ({\color{blue} blue solid line~---}) and
EWA (black dotted line - - -).}
\end{figure}

\begin{figure}
\begin{center}
\includegraphics[width=\textwidth]{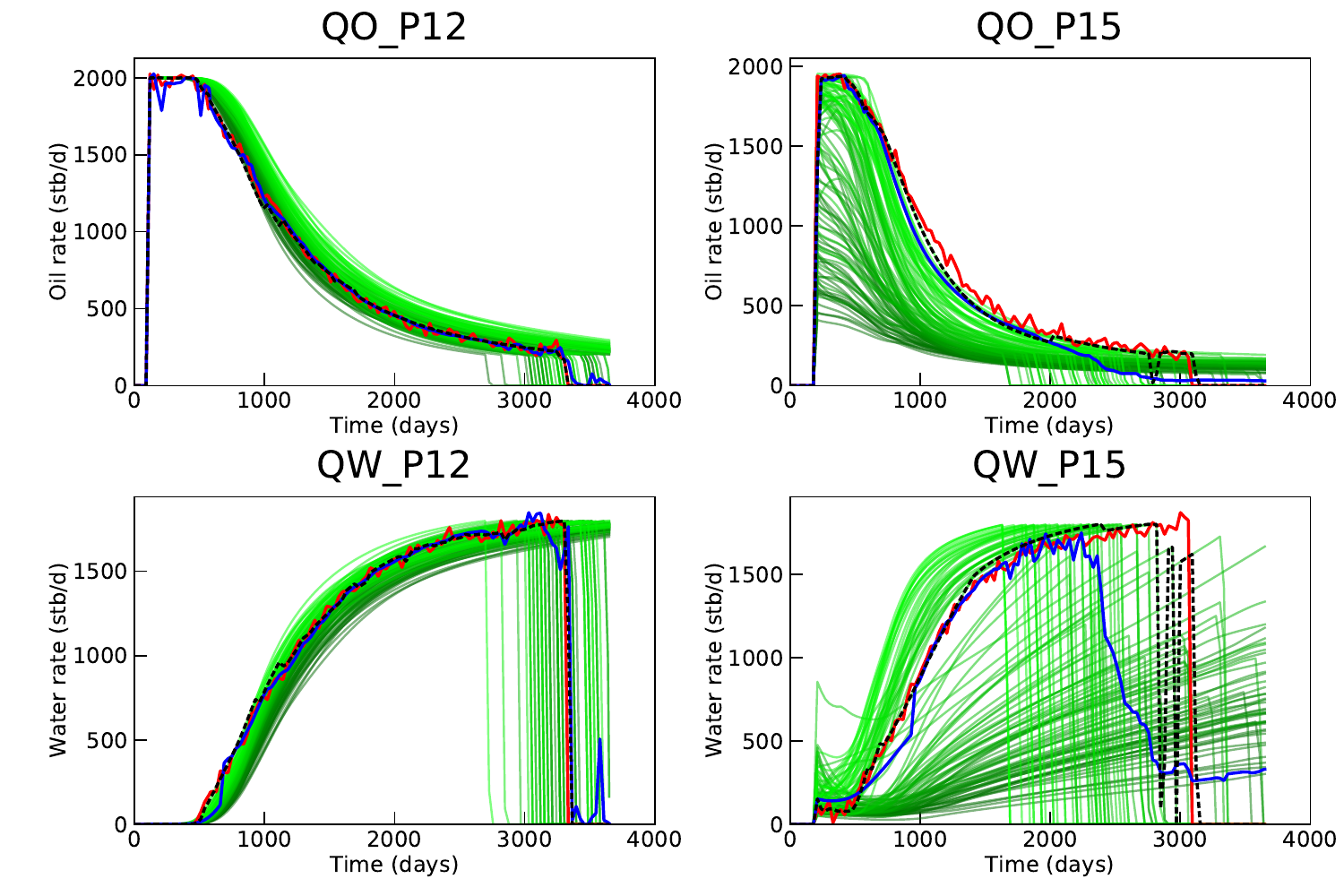}
\end{center}
\caption{\label{fig:RMSE-qualiD} Values simulated for the 104 models ({\color{green} green lines ---}),
observations ({\color{red} red solid line~---}),
one-step-ahead
forecasts by Ridge ({\color{blue} blue solid line~---}) and
EWA (black dotted line - - -).}
\end{figure}

Figures~\ref{fig:RMSE-quali} and~\ref{fig:RMSE-qualiD} report the forecasts of Ridge and EWA
for 12 time-series, considered representative of the 70 time-series
to be predicted.

The main comment would be that the aggregated forecasts look generally close enough to the
observations, even though most of the model forecasts they build on
may err. The aggregated forecasts evolve typically in a smoother way than
the observations. The right-most part of the BHP\_I1 picture reveals that Ridge is typically less
constrained than EWA by the ensemble of forecasts: while EWA cannot
provide aggregated forecasts that are out of the convex hull of the ensemble
forecasts, Ridge resorts to linear combinations of the latter and may thus
output predictions out of this range.

Also, Ridge typically adjusts faster to regime changes, while EWA takes some more time to depart from a simulation it had stuck to. This can lead to better results for Ridge, as can be seen for instance for BHP\_I8, BHP\_P12 or QO\_P19. However, it may at times react too fast and be less successful than EWA in predicting discontinuous changes in the production curves, typically pressure drop, water breakthrough or well shut-in. This is visible in Figure~\ref{fig:RMSE-quali} for pressure at well P12 in the initial time steps (bumps), for water breakthrough time at wells P9 and P18, and in Figure~\ref{fig:RMSE-qualiD} for the well closures due to the water-cut constraint at the producers. EWA appears thus more reliable than Ridge. However, its performance is very dependent on the choice of the base models as illustrated in Figure~\ref{fig:RMSE-qualiD} for property QW\_P15: the observed data being outside the range of simulated values at the end of the production period, the aggregated forecast can in no way be predictive.

Figure~\ref{fig:Lasso-P1} shows the aggregated forecasts obtained with Lasso for 4 time-series. The results seem globally better than with the two other approaches, even if, as for Ridge, we can see bumps at the beginning of the production period for BHP\_P12 and QW\_P18.

Due to the nature of the discontinuous events occurring in this study, the time they take place is not known in advance, making them difficult to predict. For events that can be included in the production schedule considered for simulating the ensemble forecasts, such as well shut-in and re-opening for testing or workover, the time of the discontinuity will be the same for the simulations and the observations. In this case, the aggregated forecasts at the time of the re-opening could rather be computed based on the misfit between the observations and the simulations prior to the shut-in.

\subsection{Quantitative study}
\label{sec:RMSE}

A more objective assessment of the performance of the approaches followed
can be obtained through an accuracy criterion. As is classical in
the literature, we resort to the root mean-square error (RMSE),
which we define by taking into account a training period:
by taking out of the evaluation the first $31$ predictions
(that is, roughly $1/4$ of the observations, $\sim 900$ days).
Hence, the algorithms are only evaluated on time steps~$32$
to~$127$, as follows:
\[
\sqrt{\frac{1}{127-32+1} \sum_{t=32}^{127} \bigl( \hat{y}_t - y_t \bigr)^2}\,.
\]

Similar formulae determine on
this time interval $32$--$127$ the performance of the best model
and of the best convex combination of the models:
\begin{align*}
\min_{j=1,\ldots,104} & \ \
\sqrt{\frac{1}{127-32+1} \sum_{t=32}^{127} ( m_{j,t} - y_t )^2} \\
\qquad \mbox{and} \qquad
\min_{(v_1,\ldots,v_{104}) \in \cC} & \ \
\sqrt{\frac{1}{127-32+1} \sum_{t=32}^{127}
\left( \sum_{j=1}^{104} v_j \, m_{j,t} - y_t \right)^{\!\! 2}}
\end{align*}
where $\cC$ denotes the set of all convex weights, i.e.,
all vectors of $\R^{104}$ with non-negative coefficients summing up to~$1$.
The best model and the best convex combination of the models
vary by the time-series; this is why we will sometimes write
the ``best local model'' or the ``best local convex combination''.

Note that the orders of magnitude of the time-series are extremely
different, depending on what is measured (they tend to be similar within wells of the same type
for a given property). We did not correct for that and did not try
to normalize the RMSEs. (Considering other criteria like the mean
absolute percentage of error -- MAPE -- would help to get
such a normalization.)

\begin{figure}
\begin{center}
Performance summary for Ridge \medskip \\
\includegraphics[width=\textwidth]{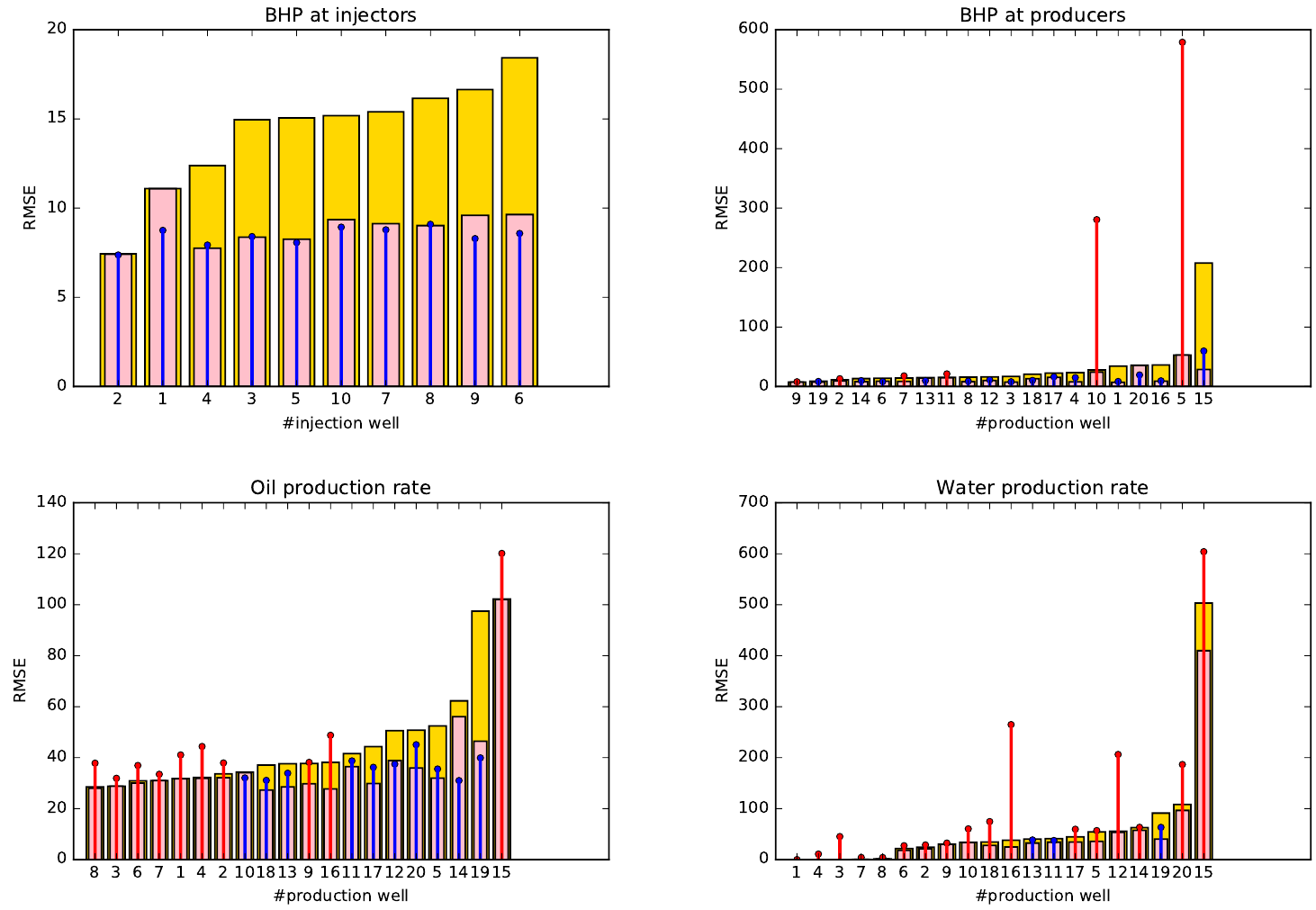}
\\ \ \\
\rule{\linewidth}{0.8pt}
\\ \ \\
Performance summary for EWA \medskip \\
\includegraphics[width=\textwidth]{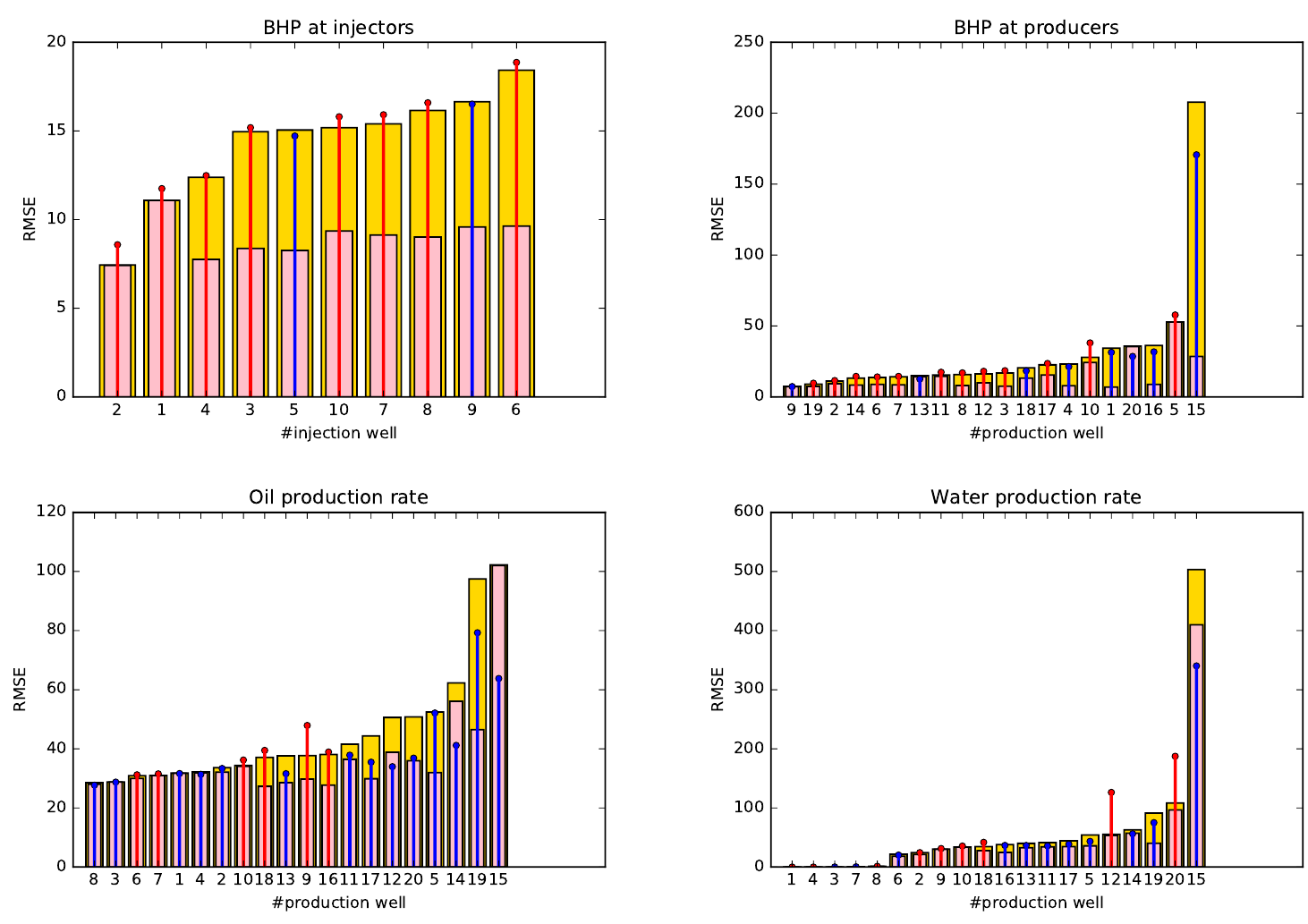}
\end{center}
\caption{\label{fig:RMSE-quanti} RMSEs of the best model
({\color{strongyellow} yellow bars}) and of the best convex combination
of models ({\color{pink} pink bars}) for each property,
as well as the RMSEs of the considered algorithms:
Ridge (top graphs) and EWA (bottom graphs).
The RMSEs of the latter are depicted in {\color{blue} blue} whenever
they are smaller than that of the best model for the considered time-series,
in {\color{red} red} otherwise.}
\end{figure}

\begin{figure}
\begin{center}
\includegraphics[width=\textwidth]{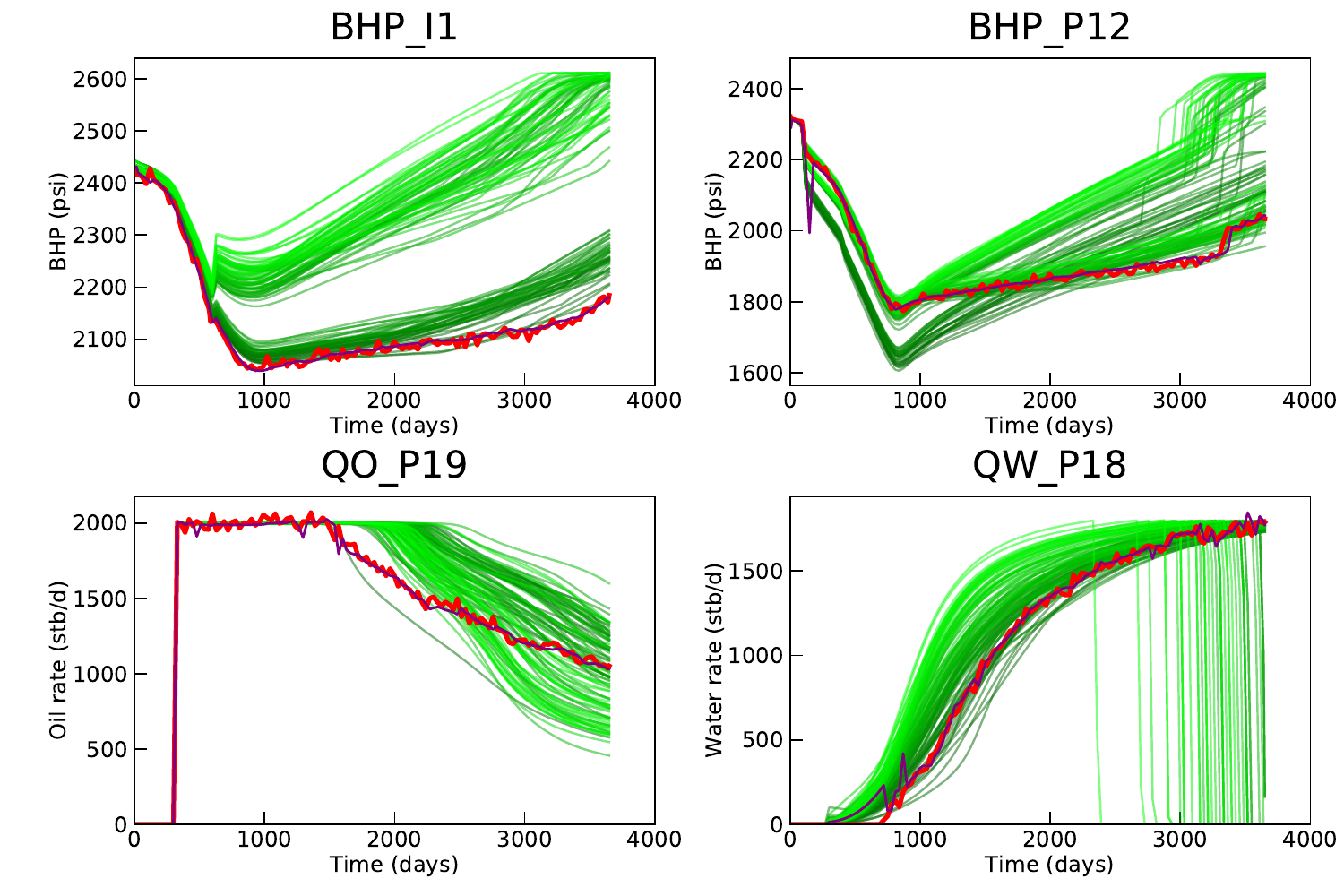}
\end{center}
\caption{\label{fig:Lasso-P1}
Values simulated for the 104 models ({\color{green} green lines ---}),
observations ({\color{red} red solid line~---}) and
one-step-ahead
forecasts by Lasso ({\color{purp} purple solid line ---}).
}
\ \\
\begin{center}
\includegraphics[width=\textwidth]{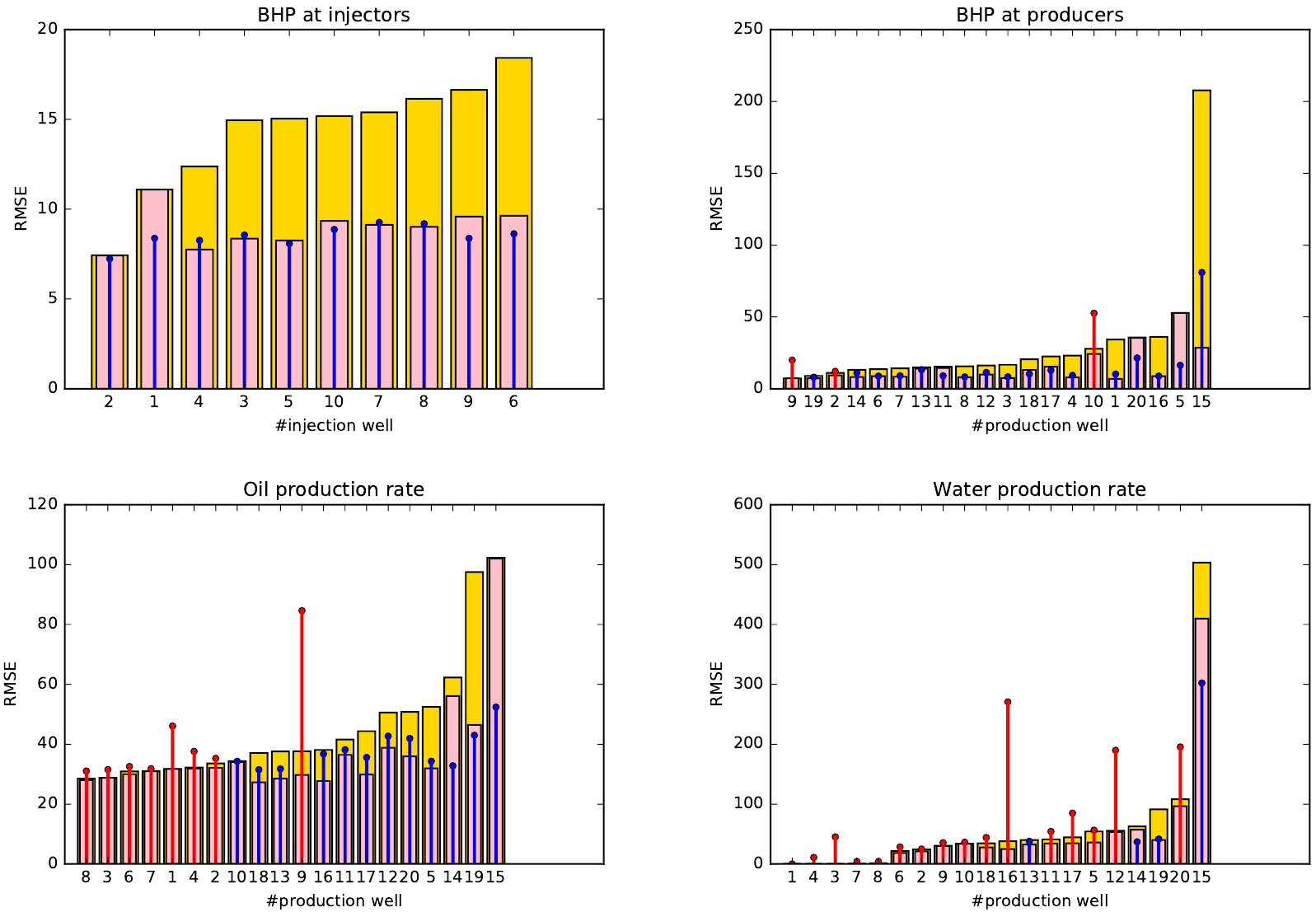}
\end{center}
\caption{\label{fig:Lasso-P2}
RMSEs of the best model
({\color{strongyellow} yellow bars}) and of the best convex combination
of models ({\color{pink} pink bars}) for each property,
as well as the RMSEs of Lasso.
The latter are depicted in {\color{blue} blue} whenever
they are smaller than that of the best model for the considered property,
in {\color{red} red} otherwise.
}
\end{figure}

The various RMSEs introduced above are represented in Figures~\ref{fig:RMSE-quanti} and~\ref{fig:Lasso-P2}.
A summary of performance would be that Ridge typically
gets an overall accuracy close to that of the best local convex
combination while EWA rather performs like the best local model. This is perfectly in
line with the theoretical guarantees described in Section~\ref{sec:appalgo}.
But Ridge has a drawback: the instabilities (the reactions that might come too fast)
already underlined in our qualitative assessment result in a few absolutely
disastrous performance, in particular for
BHP\_P5, BHP\_P10, QW\_P16, QW\_P12. The EWA algorithm seems a safer
option, though not being as effective as Ridge. The deep reason why EWA is safer
comes from its definition: it only resorts to convex weights of the model forecasts,
and never predicts a value larger (smaller) than the largest (smallest)
forecast of the models.

As can be seen in Figure \ref{fig:Lasso-P2}, the accuracy achieved by Lasso is slightly better than that of Ridge,
with only one exception, the oil production rate at well P9.
Otherwise, Lasso basically gets the best out of the accuracy of EWA
(which predicts globally well all properties for producers, namely, bottomhole pressure, oil production rate and
water production rate) and that of Ridge (which predicts well the bottomhole pressure
for injectors). However, this approach does not come with any theoretical guarantee so far.
We refer the interested reader to~\cite{Desw17} for an illustration of the
selection power of Lasso (the fact that the weights it outputs are sparse, i.e., that it discards
many simulated values).

\section{Results for multi-step-ahead forecasts}
\label{sec:faisceaux}

In this section, the simulation period of 10 years is divided into two parts: the first two thirds (time steps 1 to 84, i.e., until $\sim 2400$ days) is considered the learning period, for which observations are assumed known; the last third corresponds to the prediction period for which we aim to provide longer-term forecasts in the form of interval forecasts as explained in Section~\ref{sec:highlevel2}.
We only discuss here the use of Ridge and EWA for which such longer-term forecasting methodologies have been provided.
We actually consider a practical twist on the implementation of Ridge (not on EWA).

\subsection{Selection of a subset of the models for Ridge}

As far as Ridge is concerned (not EWA), we actually do not use
all the models in the prediction part of the data set, but only
the most reasonable ones: the ones whose root mean-square error on
the learning part of the data set is smaller than 10 times the one
of the best model. The forecasts of these models are printed in green
in Figures~\ref{fig:Good-Ridge} and~\ref{fig:Bad-Ridge}
while the forecasts of the models discarded due to this rule are in grey.
We did so for the sake of practical performance and numerical stability.

\subsection{Some good results}

Figures~\ref{fig:Good-Ridge} and~\ref{fig:Good-EWA}
report interval forecasts that look good: they are significantly
narrower than the sets of scenarios while containing most of the observations.

They were obtained, though, by using some \emph{hand-picked} parameters~$\lambda$ or~$\eta$:
we manually performed some trade-off between the widths of the interval forecasts (which is expected to be much
 smaller than the set $S$ of all scenarios) and the accuracy of the predictions (a large proportion of future observations should lie in the interval forecasts).

We were unable so far to get any satisfactory automatic tuning of these parameters
(unlike the procedure that we described in Section~\ref{sec:implement} for one-step-ahead prediction).
Hence, the good results achieved on Figures~\ref{fig:Good-Ridge} and~\ref{fig:Good-EWA} merely hint
at the potential benefits of our methods once they will come with proper parameter-tuning rules.

\newcommand{\CommRidge}{
Values simulated for the 104 models ({\color{green} green lines ---}
or {\color{grey} grey lines ---}, depending on whether the simulations were selected for
the interval forecasts), observations ({\color{red} red solid line~---}),
set $S$ of scenarios (upper and lower bounds given by black dotted lines - - -),
and interval forecasts output by Ridge (upper and lower bounds given by
{\color{blue} blue solid lines ---}). Values of~$\lambda$ used are
written on the graphs. The grey vertical line denotes the beginning of the prediction period.
}

\newcommand{\CommEWA}{
Values simulated for the 104 models ({\color{green} green lines ---}),
observations ({\color{red} red solid line~---}),
set $S$ of scenarios (upper and lower bounds given by black dotted lines - - -),
and interval forecasts output by EWA (upper and lower bounds given by
\textbf{solid lines ---}). Values of $\eta$ used are
written on the graphs. The grey vertical line denotes the beginning of the prediction period.
}

\begin{figure}[p!h!]
\begin{center}
\includegraphics[width=\textwidth]{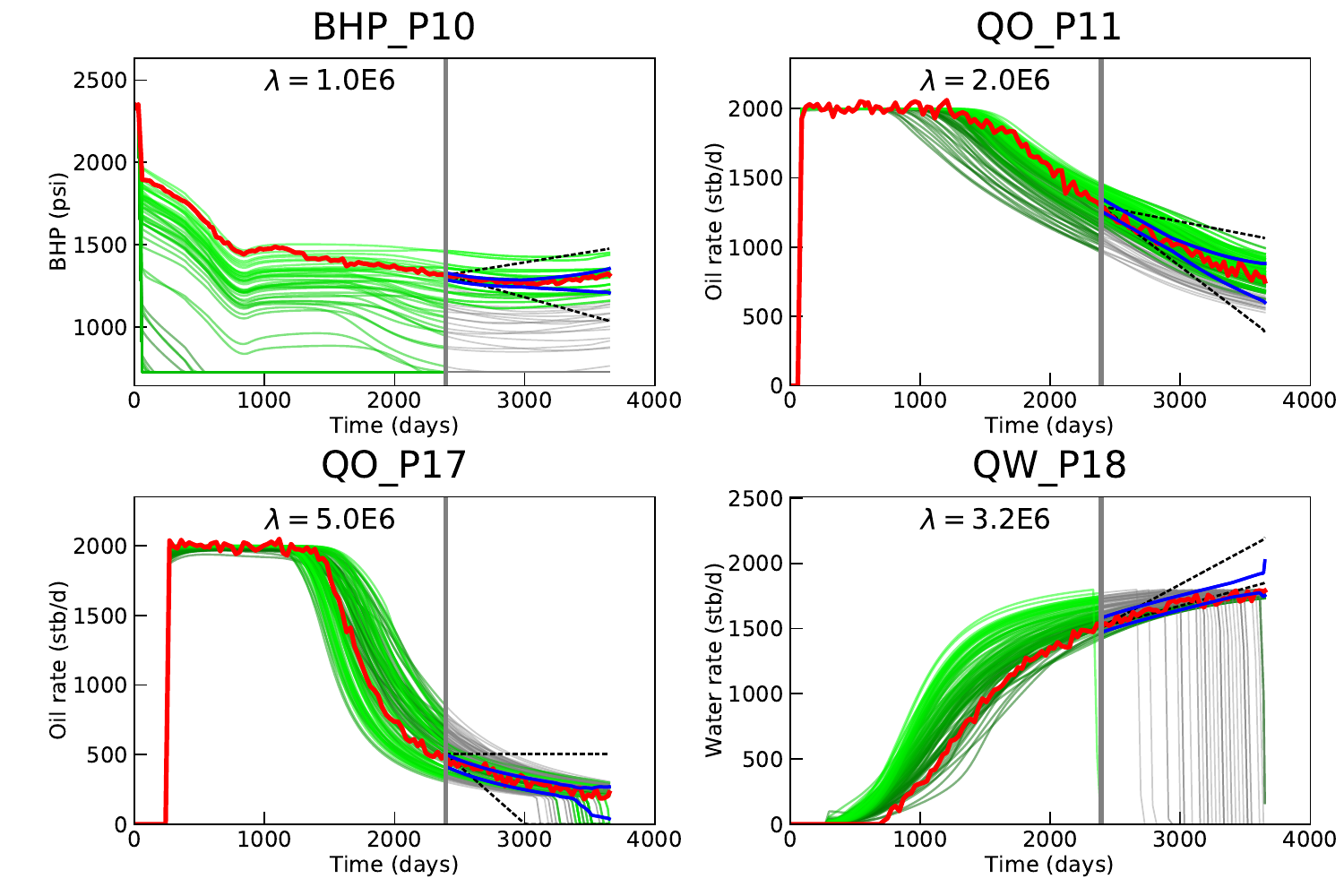}
\end{center}
\caption{
\label{fig:Good-Ridge}
\CommRidge \vspace{.2cm}
}
\begin{center}
\includegraphics[width=\textwidth]{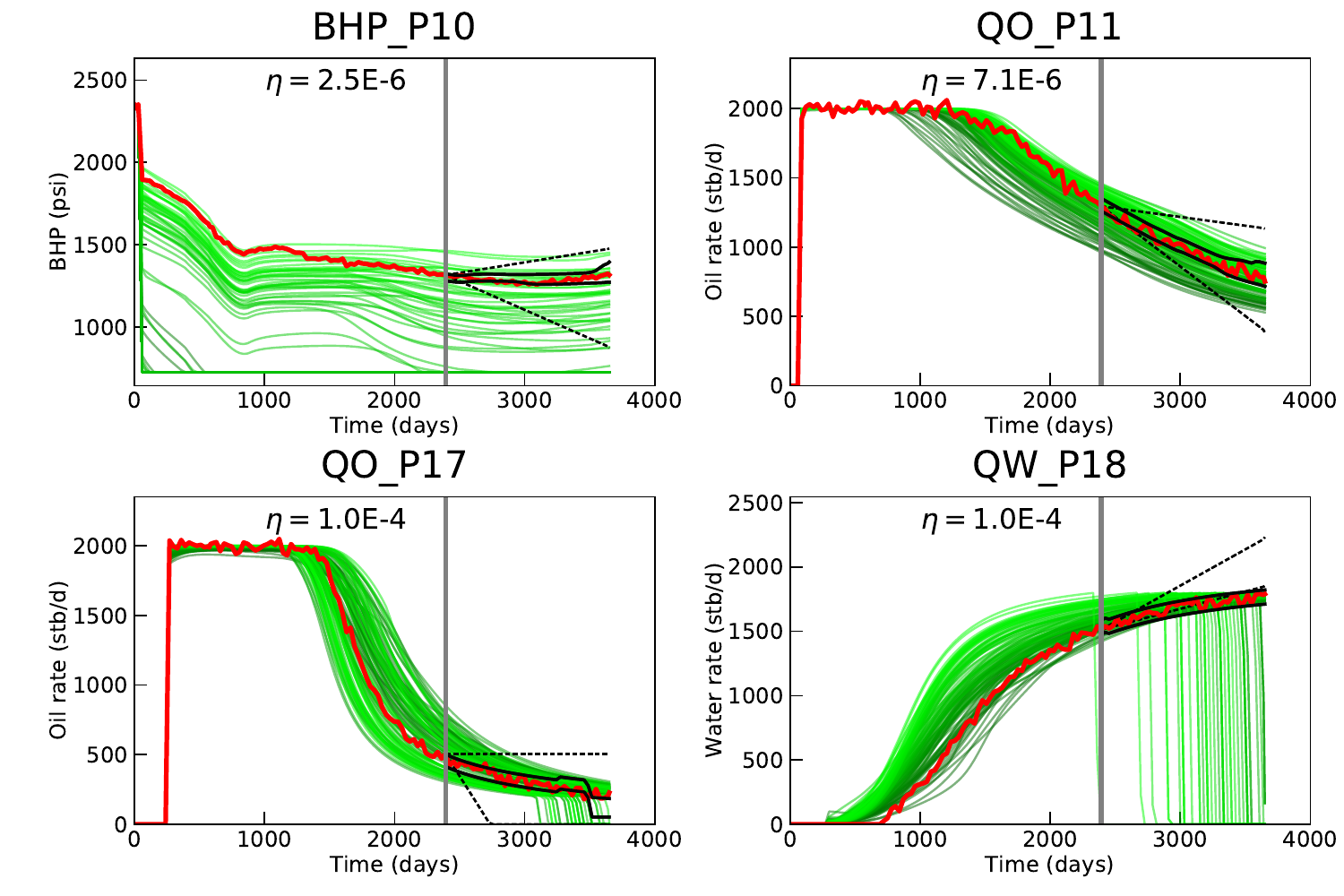}
\end{center}
\caption{
\label{fig:Good-EWA}
\CommEWA
}
\end{figure}

\subsection{Some disappointing results}

\begin{figure}
\begin{center}
\includegraphics[width=\textwidth]{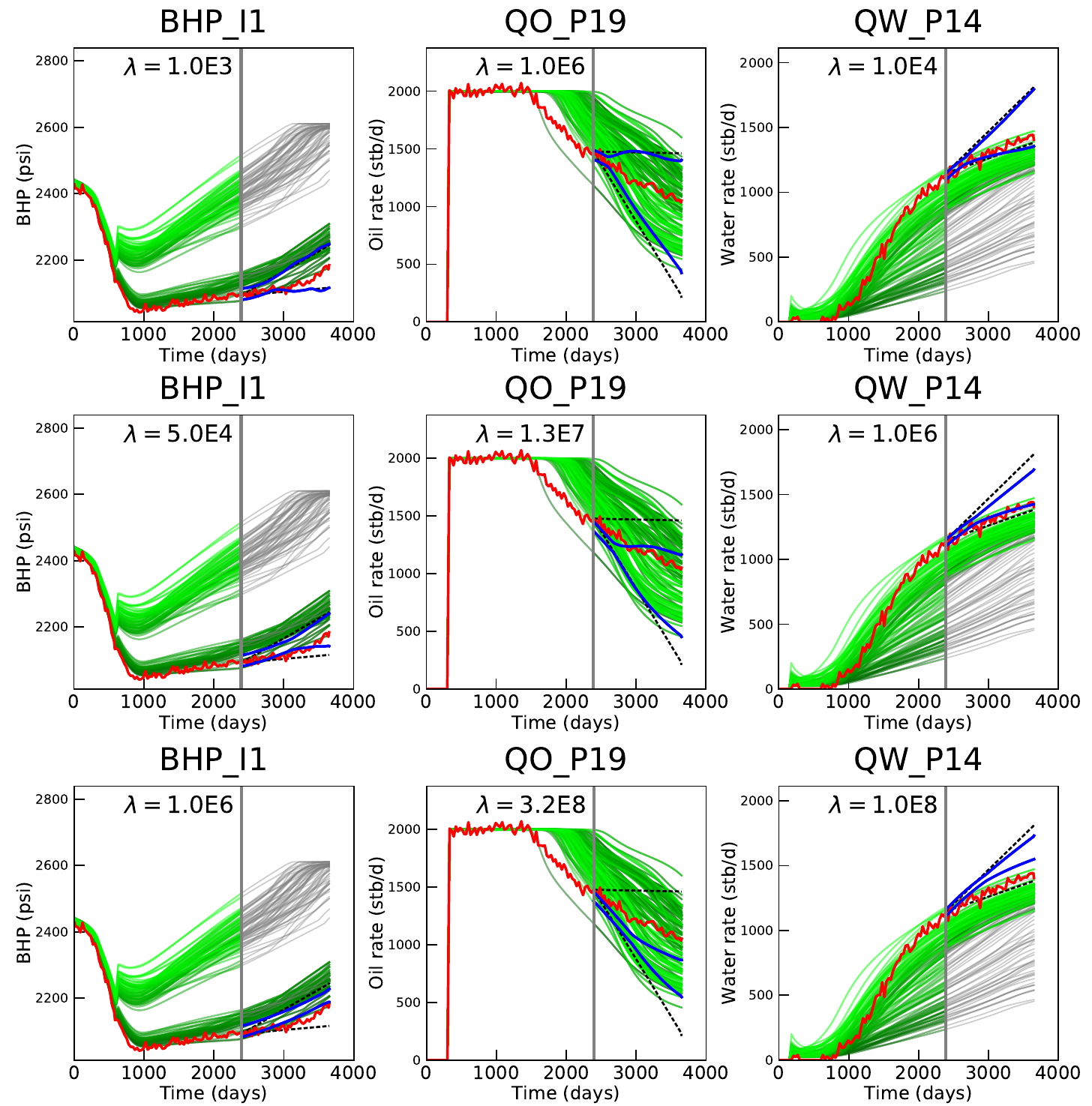}
\end{center}
\caption{\label{fig:Bad-Ridge} \CommRidge}
\end{figure}

\begin{figure}
\begin{center}
\includegraphics[width=\textwidth]{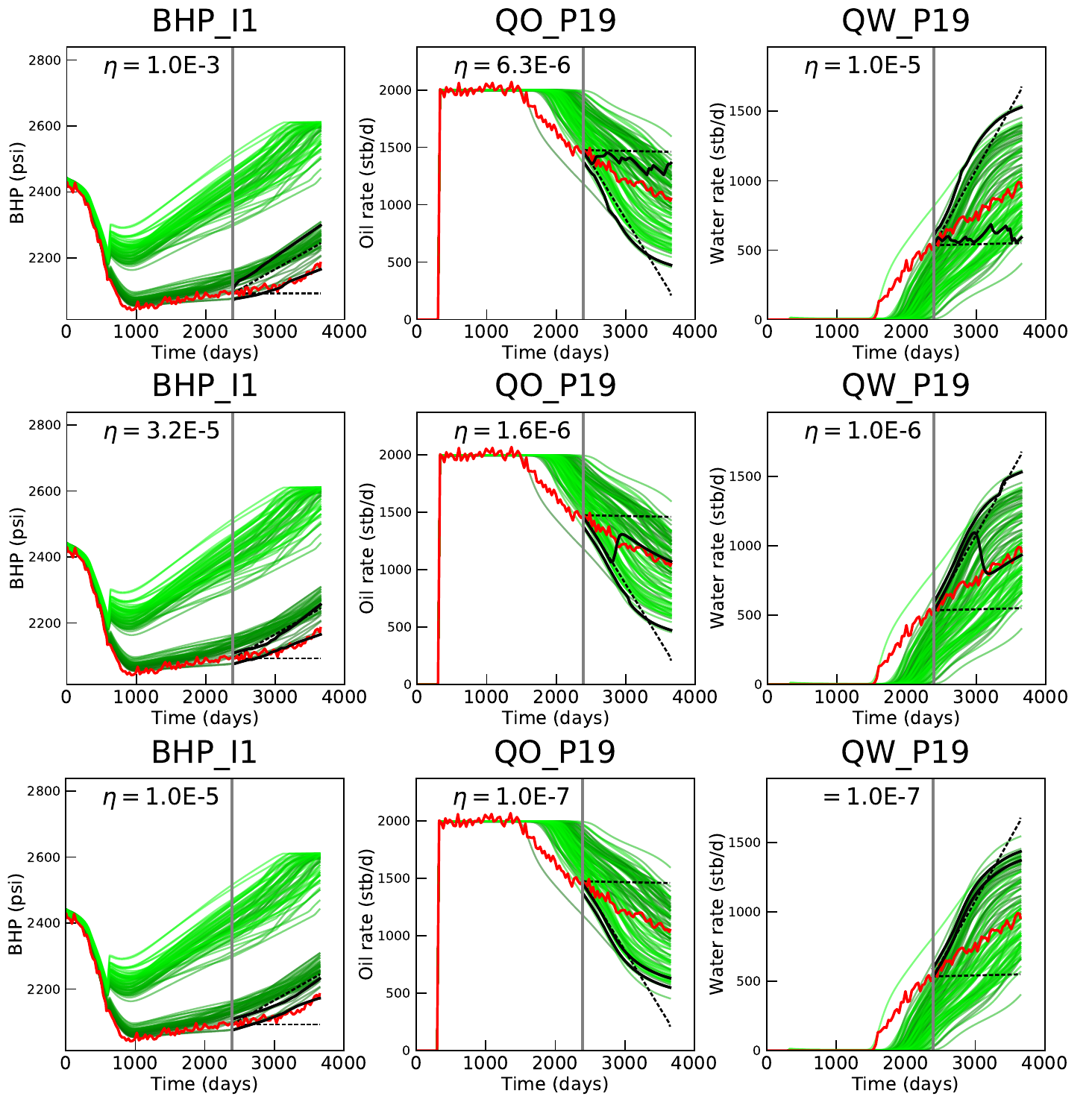}
\end{center}
\caption{\label{fig:Bad-EWA} \CommEWA}
\end{figure}

Figures~\ref{fig:Bad-Ridge} and~\ref{fig:Bad-EWA} show, on the other
hand, that for some time-series, neither Ridge nor EWA
may provide useful interval forecasts: the latter either
completely fail to accurately predict the observations
or they are so large that they cover (almost) the set of
all scenarios -- hence, they do not provide any useful
information.

We illustrate this by letting $\lambda$
increase (Figure~\ref{fig:Bad-Ridge}) and
$\eta$ decrease (Figure~\ref{fig:Bad-EWA}):
the interval forecasts become narrower as the parameters vary in this way.
They first provide intervals (almost) covering all scenarios and finally resort to inaccurate interval forecasts.
This behavior is the one
expected by a theoretical analysis: as the learning rate $\eta$ tends to~$0$
or the regularization parameter~$\lambda$ tends to $+\infty$, the
corresponding aggregated forecasts of EWA and Ridge tend to
the uniform mean of the simulated values and 0, respectively. In particular,
variability is reduced. In constrast, when $\eta$ is large and $\lambda$ is small,
past values, including the plausible continuations $z_{T+k}$ discussed in Section~\ref{sec:highlevel2},
play an important role for determining the weights. The latter vary much as, in particular,
the plausible continuations are highly varied. \\

For time-series BHP\_I1, we can see that the interval forecast is satisfying at the beginning of the prediction period (until $\sim 3000$ days, see Figures~\ref{fig:Bad-Ridge} and~\ref{fig:Bad-EWA}), but then starts to diverge from the data. In this second period, the data are indeed quite different from the forecasts, and lie outside of the simulation ensemble. Similarly, observations for QO\_P19 and QW\_P14 show trends globally different from the simulated forecasts. This may explain the disappointing results here. Indeed, the approach appears highly dependent on the base forecasts, especially the ones that perform well in the learning period. If the latter diverge from the true observations in the prediction period, the aggregation may fail to correctly predict the future behavior.

In the case study considered here, we used the 104 geological models provided to the benchmark participants for aggregation on the entire production period. In practice, we could consider updating the ensemble of base models when the simulated forecasts start to diverge from the data, as is the case for well P19. This requires generating a new ensemble of geostatistical realizations for the petrophysical properties that is more suitable in some sense. More generally, the choice of the base ensemble should be investigated in more details in future work to improve the applicability of the proposed aggregation approaches.

\section{Conclusions and perspectives}

In this paper, we investigated the use of deterministic aggregation algorithms to predict the future production of a reservoir based on past production data. These approaches consist in aggregating a set of base simulated forecasts to predict production at the next time step. To answer the need for longer-term predictions in reservoir engineering, an extension of these approaches was developed for multi-step-ahead predictions that provides interval forecasts based on some putative observations in the prediction period.

These approaches were applied on the Brugge test case. The one-step-ahead forecasts were globally satisfactory, even very satisfactory. For longer-term predictions, the proposed approaches yield mixed results. Sometimes they can lead to accurate and narrow interval forecasts, and sometimes they
can fail in providing an interval forecast that is narrower than the initial putative interval or that contains most of the future observations. This may be due to the lack of relevance of the set of base forecasts, not informative enough as far as future observations are concerned. More research is needed
to find automatic ways to correct inaccurate sets of base forecasts. \\

In the future, and once the methodology for longer-term predictions is fully developed, it would be interesting to test the proposed approaches on other reservoir case studies. New developments could also be envisioned, such as the extension of Lasso to multi-step-ahead predictions. Another interesting work would be to identify quantitative criteria to validate the interval forecasts output by the aggregation techniques. Finally, the aggregation was performed here time-series by time-series, meaning one property and one well at a time. In the future, it would be interesting to consider a simultaneous aggregation of all time-series, with identical weights for all of them. This could pave the way to other applications, such as the prediction of the dynamic behavior at a new well, or a better characterization of the model parameters.

\vspace{1cm}

\section*{Acknowledgments}
The authors would like to thank Charles-Pierre Astolfi who worked as an M.Sc.\ intern
on a preliminary data set linked to this project, for one-step-ahead predictions.

\newpage
\bibliographystyle{spmpsci}      
\bibliography{Bib}

\end{document}